\documentclass[twoside,11pt]{article}
\usepackage{amsmath}
\usepackage{jmlr2e}
\usepackage{booktabs}       
\usepackage{algorithm}
\usepackage[noend]{algpseudocode}

\usepackage{booktabs}
\usepackage{threeparttable}

\newtheorem{assumption}{Assumption}

\firstpageno{1}

\begin{document}

\title{
Graph Interpolating Activation Improves Both Natural and Robust Accuracies in Data-Efficient Deep Learning
}

\author{\name Bao Wang \email wangbaonj@gmail.com \\
       \addr Department of Mathematics\\
       University of California, Los Angeles\\
       Los Angeles, CA 90095-1555, USA
       \AND
       \name Stanley J.\ Osher \email sjo@math.ucla.edu \\
       \addr Department of Mathematics\\
       University of California, Los Angeles\\
       Los Angeles, CA 90095-1555, USA}

\editor{} 

\maketitle

\begin{abstract}
Improving the accuracy and robustness of deep neural nets (DNNs) and adapting them to small training data are primary tasks in deep learning research. In this paper, we replace the output activation function of DNNs, typically the data-agnostic softmax function, with a graph Laplacian-based high dimensional interpolating function which, in the continuum limit, converges to the solution of a Laplace-Beltrami equation on a high dimensional manifold. Furthermore, we propose end-to-end training and testing algorithms for this new architecture. The proposed DNN with graph interpolating activation integrates the advantages of both deep learning and manifold learning. Compared to the conventional DNNs with the softmax function as output activation, the new framework demonstrates the following major advantages: First, it is better applicable to data-efficient learning in which we train high capacity DNNs without using a large number of training data. Second, it remarkably improves both natural accuracy on the clean images and robust accuracy on the adversarial images crafted by both white-box and black-box adversarial attacks. Third, it is a natural choice for semi-supervised learning. 
For reproducibility, the code is available at \url{https://github.com/BaoWangMath/DNN-DataDependentActivation}.
\end{abstract}


\section{Introduction}
Deep learning (DL) has achieved tremendous success in both image and speech recognition and natural language processing, and it has been widely used in industrial production \citep{DeepLearningReview:2015}. Improving generalization accuracy and adversarial robustness of deep neural nets (DNNs) are primary tasks in DL research. Moreover, applying DNNs to data-efficient machine learning (ML), where we do not have a large number of training instances, is important to different research communities.
\medskip

\noindent Despite the extraordinary success of DNNs in image and speech perception, their vulnerability to adversarial attacks raises concerns when applying them to security-critical tasks, e.g., autonomous cars, robotics, and DNN-based malware detection systems \citep{Attack:Tesla}. Since the seminal work of \cite{szegedy2013intriguing}, recent research shows that DNNs are vulnerable to many kinds of adversarial attacks including physical, poisoning, and inference (evasion) attacks \citep{Chen:2017,CWAttack:2016,PapernotAttack:2016,Goodfellow:2014AdversarialTraining}. Physical attacks occur during data acquisition, poisoning and inference attacks happen during training and testing phases of machine learning (ML), respectively.
\medskip

\noindent Adversarial attacks have been successful in both white-box and black-box scenarios. In white-box attacks the adversarial have access to the architecture and weights of DNNs. In black-box attacks the adversarial have no access to the details of the underlying model. Black-box attacks are successful because one can perturb an image to cause its misclassification on one DNN, and the same perturbed image also has a significant chance to be misclassified by another DNN; this is known as the transferability of adversarial examples \citep{DBLP:journals/corr/PapernotMG16}. Due to this transferability, it is straightforward to attack DNNs in a black-box fashion by attacking an oracle model \citep{LiuYanpei:2016,Brendel:2017}. There also exist universal perturbations that can imperceptibly perturb any image and cause misclassification for any given network \citep{Moosavi-Dezfooli_2017_CVPR}. \cite{Dou:2018} analyzed the efficiency of many adversarial attacks for a large variety of DNNs.
\medskip

\noindent Besides the issue of adversarial vulnerability, the superior accuracy of DNNs depends heavily on a massive amount of training data. When we do not have sufficient training data, which is often the case in many real situations, to train a high capacity deep network, performance degradation becomes a serious problem. As shown in Table~\ref{Table:Cifar10-SmallSet}, when ResNets are trained on $50$K or $10$K CIFAR10 images, as the depth of ResNet increases, the test accuracy gains. However, when ResNets are trained on only $1$K images, the test accuracy decays as the model's capacity increases. For instance, the test errors of ResNet20 and ResNet110 are $34.90$\% and $42.94$\%, respectively.

\begin{table*}[!htbp]
\centering
\caption{Test errors of DNNs trained on the entire (50K), the first 10K, and the first 1K instances of the training set of the CIFAR10.}
\label{Table:Cifar10-SmallSet}
\begin{tabular}{ccccc}
\toprule[1.0pt]
Network & \# of parameters &  50K  & 10K & 1K \\
\midrule[0.8pt]
ResNet20 &	0.27M	&9.06\% (8.75\%\citep{ResNet})  &12.83\%  & 34.90\%  \\
ResNet32 &	0.46M	&7.99\% (7.51\%\citep{ResNet})  &11.18\%  & 33.41\%  \\
ResNet44 &	0.66M	&7.31\% (7.17\%\citep{ResNet})  &10.66\%  & 34.58\%  \\
ResNet56 &	0.85M	&7.24\% (6.97\%\citep{ResNet})  &9.83\%   & 37.83\%  \\
ResNet110& 	1.7M 	&6.41\% (6.43\%\citep{ResNet})  &8.91\%   & 42.94\%  \\
\bottomrule[1.0pt]
\end{tabular}
\end{table*}

\subsection{Our Contributions}
\noindent In this paper, we propose an end-to-end framework to mitigate the aforementioned two issues of DNNs, i.e., adversarial vulnerability and generalization accuracy degradation in the small training data scenario. At the core of our framework is to replace the data-agnostic softmax output activation with a data-dependent graph interpolating function. To this end, we leverage the weighted nonlocal Laplacian (WNLL) \citep{Shi:2018WNLL} to interpolate features in the hidden state of DNNs. In back-propagation, we linearize the WNLL activation function to compute gradient of the loss function approximately. The major advantages of the proposed framework are summarized below.

\begin{itemize}
    \item The naturally trained DNNs with the WNLL output activation obtained by solving the empirical risk minimization (ERM), i.e., Eq.~(\ref{ERM}), are remarkably more accurate than the vanilla DNNs with the softmax output activation. 
    
    \item The robustly trained DNNs with the WNLL activation obtained by solving the empirical adversarial risk minimization (EARM), i.e., Eq.~(\ref{Adversarial-ERM}), are much more robust to adversarial attacks than the robustly trained vanilla DNNs. To the best of our knowledge, DNNs with the WNLL activation achieves the current-state-of-the-art result in adversarial defense on the CIFAR10 and MNIST benchmarks.
    
    \item In the small training data situation, the WNLL activation can regularize the training procedure. The test accuracy of DNNs with the WNLL activation increases as the network goes deeper.
    
    \item DNN with the WNLL output activation is a natural choice for semi-supervised deep learning.
    
    \item The proposed framework is applicable to any off-the-shelf DNNs when use the softmax as its output activation.
\end{itemize}

\subsection{Related Work}
In this subsection, we will discuss related work from the viewpoints of improving generalizability and adversarially robustness.

\subsubsection{Improving Generalizability of DNNs}
Generalizability is crucial to DL, and many efforts have been made to improve the test accuracy of DNNs \citep{GreedyTraining:2007,DBN:2006}. Advances in network architectures such as VGG networks \citep{VGG:2014}, deep residual networks (ResNets) \citep{ResNet,PreActResNet} and recently DenseNets \citep{DenseNet} and many others \citep{Chen:2017NIPS}, together with powerful hardware make the training of very deep networks with good generalization capabilities possible. Effective regularization techniques such as dropout and maxout \citep{hinton2012improving, wan2013regularization, goodfellow2013maxout}, as well as data augmentation methods \citep{krizhevsky2012imagenet,VGG:2014} have also explicitly improved generalization for DNNs. From the optimization point of view, Laplacian smoothing stochastic gradient descent has been recently proposed to improve training and generalization of DNNs \citep{LS-GD:2018}.
\medskip

\noindent A key component of DNN is the activation function. Improvements in designing of activation functions such as the rectified linear unit (ReLU) \citep{glorot2011deep}, have led to huge improvements in performance in computer vision tasks \citep{nair2010rectified, krizhevsky2012imagenet}. More recently, activation functions adaptively trained to the data such as the adaptive piece-wise linear unit (APLU) \citep{agostinelli2014learning} and parametric rectified linear unit (PReLU) \citep{he2015delving} have led to further improvements in the performance of DNNs. For output activation, support vector machine (SVM) has also been successfully applied in place of softmax \citep{Tang:2013}. Though training DNNs with softmax or SVM as output activation is effective in many tasks, it is possible that alternative activations that consider the manifold structure of data by interpolating the output based on both training and testing data can boost the performance of the deep network. In particular, ResNets can be modeled as solving control problems of a class of transport equations in the continuum limit \citep{ResNet:PDE,EnResNet:Wang}. Transport equation theory suggests that using an interpolating function that interpolates terminal values from initial values can dramatically simplify the control problem compared with an ad-hoc choice. This further suggests that a fixed and data-agnostic activation for the output layer may be suboptimal.

\subsubsection{Adversarial Defense}
EARM is one of the most successful mathematical frameworks for certified adversarial defense. Under the EARM framework, adversarial defense for the
$\ell_\infty$-norm based inference attacks can be formulated as solving the following minimax optimization problem 
\begin{equation}
\label{Adversarial-ERM}
\min_{f\in \mathcal{H}} \frac{1}{n} \sum_{i=1}^n \max_{\|\mathbf{x}'_i-\mathbf{x}_i\|_\infty \leq \epsilon} L(f(\mathbf{x}_i', \mathbf{w}), y_i),
\end{equation}
where $f(\cdot, \mathbf{w})$ is a function in the hypothesis class $\mathcal{H}$, e.g., DNNs, parameterized by $\mathbf{w}$. Here, $\{(\mathbf{x}_i, y_i)\}_{i=1}^n$ are $n$ i.i.d. data-label pairs drawn from some high dimensional unknown distribution $\mathcal{D}$, $L(f(\mathbf{x}_i, \mathbf{w}), y_i)$ is the loss associated with $f$ on the data-label pair $(\mathbf{x}_i, y_i)$. For classification, $L$ is typically selected to be the cross-entropy loss; for regression, the root mean square error is commonly used. The adversarial defense for other measure based attacks can be formulated similarly. As a comparison, solving ERM is used to train models in a natural fashion to classify the clean data, where ERM is to solve the following optimization problem
\begin{equation}
\label{ERM}
\min_{f\in \mathcal{H}} \frac{1}{n} \sum_{i=1}^n L(f(\mathbf{x}_i, \mathbf{w}), y_i).
\end{equation}

\noindent Many of the existing approaches try to defend against the inference attacks by searching for a good surrogate loss to approximate the loss function in the EARM. Projected gradient descent (PGD) adversarial training is a representative work along this line that approximates EARM by replacing $\mathbf{x}_i'$ with the adversarial data that is obtained by applying the PGD attack to the clean data \citep{Goodfellow:2014AdversarialTraining,Madry:2018}. Besides finding an appropriate surrogate to approximate the empirical adversarial risk, under the EARM framework, we can also improve the hypothesis class to improve adversarial robustness of the trained robust models \citep{EnResNet:Wang}.
\medskip

\noindent There is a massive volume of research over the past several years on defending against adversarial attacks for DNNs. Randomized smoothing transforms an arbitrary classifier $f$ into a ''smoothed" surrogate classifier $g$ and is certifiably robust to the $\ell_2$-norm based adversarial attacks \citep{Lecuyer:2019,CRK19}. Among the randomized smoothing technique, one of the most popular ideas is to inject Gaussian noise to the input image, and the classification result is based on the probability of noisy image in decision region. \cite{EnResNet:Wang} modeled ResNets as a transport equation and interpreted the adversarial vulnerability of DNNs as irregularity of the transport equation's solution. To enhance its regularity, i.e., improve adversarial robustness, they added a diffusion term to the transport equation and solved the resulted convection-diffusion equation by the celebrated Feynman-Kac formula. The resulted algorithm remarkably improves both natural and robust accuracies of the robustly trained DNNs.
\medskip

\noindent Robust optimization for solving EARM has achieved tremendous success in certified adversarial defense \citep{Madry:2018,Zhang:2019-Trades}. Regularization in EARM can further boost the robustness of the adversarially trained robust models \citep{
Kurakin:2017,Ross:2017,Zheng:2017
}. The adversarial defense algorithms should learn a classifier with high test accuracy on both clean and adversarial data. To achieve this goal, \cite{Zhang:2019-Trades} developed a new loss function named TRADES that explicitly trades off between natural and robust generalization.
\medskip

\noindent Besides robust optimization, there are many other approaches for adversarial defense. Defensive distillation was proposed to increase the stability of DNN \citep{PapernotDistillationDefense:2016}, and a related approach \citep{tramer2018ensemble} cleverly modifies the training data to increase robustness against black-box attacks and adversarial attacks in general. To counter adversarial perturbations, \cite{ChuanGuo:2018} proposed to use image transformations, e.g., bit-depth reduction, JPEG compression, total variation minimization, and image quilting. These input transformations are intended to be non-differentiable, thus making adversarial attacks more difficult, especially for gradient-based attacks.
GANs are also used for adversarial defense \citep{Samangouei:2018}. However, adversarial attacks can break these gradient mask based defenses by circumventing the obfuscated gradient \citep{Athalye:2018}.
\medskip 

\noindent Instead of using the softmax function as DNN's output activation, \cite{BaoWang:2018NIPS} utilized a non-parametric graph interpolating function which provably converges to the solution of a Laplace-Beltrami equation on a high dimensional manifold \citep{Shi:2018WNLL}. 
The proposed data-dependent activation shows a remarkable amount of generalization accuracy improvement, and the results are more stable when one only has a limited amount of training data. This data-dependent activation is also useful in adversarial defense when combined with image transformations \citep{Wang:2018AdversarialDefense}.
\cite{Verma:2018} simplified the interpolation procedure and generalized it to more hidden layers to learn better representations.

\subsection{Organization}
This paper is structured in the following way: In section~\ref{Section:Architecture}, we present the generic architecture of DNNs with a graph interpolating function as its output activation. In section~\ref{Section:Algorithms}, we present training and testing algorithms in both natural and robust fashions for the proposed DNNs with graph interpolating activation. We verify the performance of the proposed algorithm numerically in section~\ref{Section:Results} from the lens of natural and robust generalization accuracies and semi-supervised learning. In section~\ref{Section:Explanation}, we provide geometric explanations for improving generalization and robustness by using the proposed new framework. This paper concludes with a remark
in section~\ref{Section:Conclusion}.

\section{Network Architecture}\label{Section:Architecture}
\begin{figure}[!ht]
\centering
\begin{tabular}{cc}
\includegraphics[width=0.48\columnwidth]{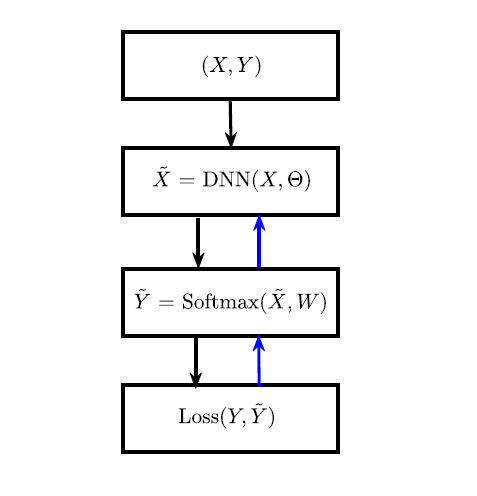}&
\includegraphics[width=0.465\columnwidth]{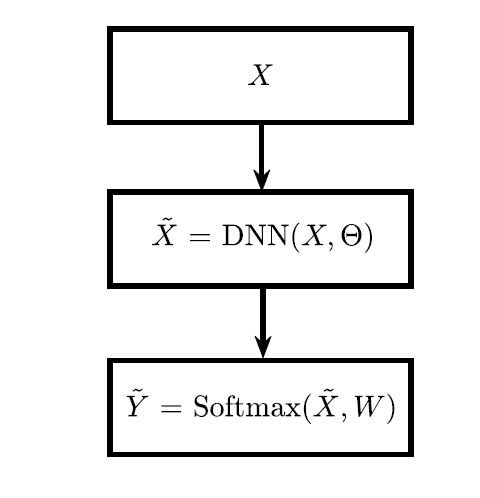}\\
(a) & (b)\\
\end{tabular}
\caption{Illustration of training and testing procedures of the standard DNN with the softmax function as output activation layer. (a): Training; (b): Testing.}
\label{fig:Standard-DNN-Structure}
\end{figure}

We illustrate the training and testing procedures of a standard DNN in Fig~\ref{fig:Standard-DNN-Structure}, where
\begin{itemize}
\item {\bf Training} (Fig~\ref{fig:Standard-DNN-Structure} (a)), in the $k$th iteration, given a mini-batch of training data $(\mathbf{X}, \mathbf{Y}$), we perform:
\begin{itemize}
    \item {\it Forward propagation:} Transform $\mathbf{X}$ into features by the DNN block (a combination of convolutional layers, nonlinearities, etc.), and then feed these features into the softmax activation to obtain the predictions $\tilde{\mathbf{Y}}$, i.e., 
$$
\tilde{\mathbf{Y}} = {\rm Softmax}({\rm DNN}(\mathbf{X}, \Theta^{k-1}), \mathbf{W}^{k-1}),
$$
where $(\Theta^{k-1}, \mathbf{W}^{k-1})$ are the temporary values of the trainable weights $(\Theta, \mathbf{W})$ at the $(k-1)$th iteration. Then the loss is computed (e.g., cross entropy) between the ground-truth labels $\mathbf{Y}$ and the predicted labels $\tilde{\mathbf{Y}}$:
$\mathcal{L} \doteq  \mathcal{L}^{\rm Linear} = {\rm Loss}(\mathbf{Y}, \tilde{\mathbf{Y}})$.

    \item {\it Backpropagation:} Update weights ($\Theta^{k-1}$, $\mathbf{W}^{k-1}$) by applying gradient descent with learning rate $\gamma$ 

$$
\mathbf{W}^{k} = \mathbf{W}^{k-1} - \gamma \frac{\partial \mathcal{L}}{\partial \tilde{\mathbf{Y}}}\cdot \frac{\partial \tilde{\mathbf{Y}}}{\partial \mathbf{W}}, 
$$
$$
\Theta^{k} = \Theta^{k-1} - \gamma \frac{\partial \mathcal{L}}{\partial \tilde{\mathbf{Y}}}\cdot \frac{\partial \tilde{\mathbf{Y}}}{\partial \tilde{\mathbf{X}}}\cdot \frac{\partial \tilde{\mathbf{X}}}{\partial \Theta}.
$$
\end{itemize}

\item {\bf Testing} (Fig~\ref{fig:Standard-DNN-Structure} (b)): Once the training procedure finishes with the learned parameters $(\Theta, \mathbf{W})$.
The predicted labels for the testing data $\mathbf{X}$ are
$$
\tilde{\mathbf{Y}} = {\rm Softmax}({\rm DNN}(\mathbf{X}, \Theta), \mathbf{W}),
$$
for notational simplicity, we still denote the test set and the learned weights as $\mathbf{X}$, $\Theta$, and $\mathbf{W}$, respectively.
\end{itemize}

\begin{figure}[!ht]
\centering
\begin{tabular}{cc}
\includegraphics[width=0.53\columnwidth]{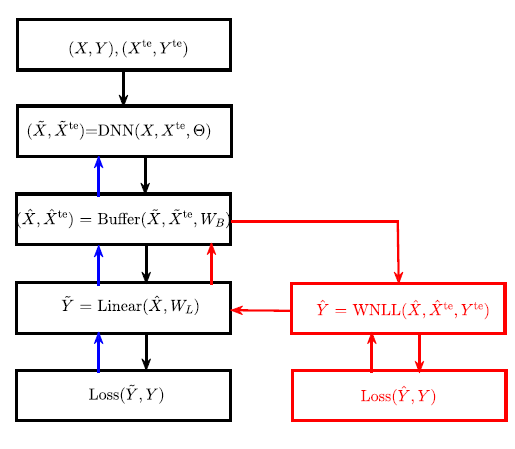}&
\includegraphics[width=0.41\columnwidth]{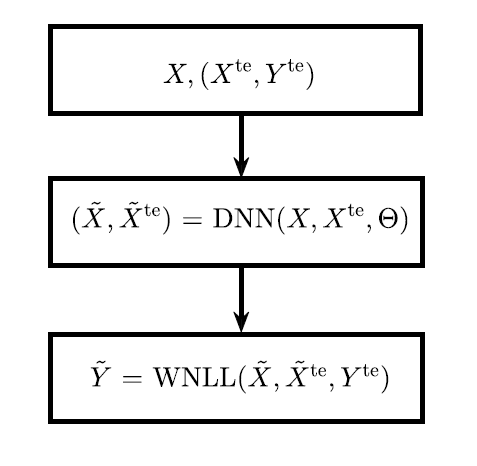}\\
(a)&(b)\\
\end{tabular}
\caption{Illustration of training and testing procedures of the DNN with the WNLL interpolating function as the output activation function. (a): Training; (b): Testing.}
\label{fig:WNLL-DNN-Structure}
\end{figure}

\noindent Even though this deep learning paradigm achieves the current-state-of-the-art success in many artificial intelligence tasks, the data-agnostic activation (softmax) acts as a linear model on the space of deep features $\tilde{\mathbf{X}}$, which does not take into consideration the underlying manifold structure of $\tilde{\mathbf{X}}$, and has many other problems, e.g., it is less applicable when we have a small amount of training data and is not robust to adversarial attacks. To this end, we replace the softmax output activation with a graph interpolating function, WNLL, which will be introduced in the following subsection. We illustrate the training and testing data flow in Fig.~\ref{fig:WNLL-DNN-Structure} which will be discussed later.

\subsection{Graph-based High Dimensional Interpolating Function -- A Harmonic Extension Approach}
Let $\mathbf{X} = \{\mathbf{x}_1, \mathbf{x}_2, \cdots, \mathbf{x}_n\}$ be a set of points located on a high dimensional manifold $\mathcal{M}\subset\mathbb{R}^d$ and $\mathbf{X}^{\rm te} = \{\mathbf{x}^{\rm te}_1, \mathbf{x}^{\rm te}_2, \cdots, \mathbf{x}^{\rm te}_m\}$ (``te" for template) be a subset of $\mathbf{X}$ that is labeled with the label function $g(\mathbf{x})$. We want to interpolate a function $u$ that is defined on the whole manifold $\mathcal{M}$ and can be used to interpolate labels for the entire dataset $\mathbf{X}$. The harmonic extension is a natural approach to find such a smooth interpolating function which is defined by minimizing the following Dirichlet energy functional
\begin{equation}
\label{DirichletEnergy}
\mathcal{E}(u)=\frac{1}{2}\sum_{\mathbf{x}, \mathbf{y}\in \mathbf{X}} w(\mathbf{x}, \mathbf{y})\left(u(\mathbf{x})-u(\mathbf{y})\right)^2,
\end{equation}
with the boundary condition
$$
u(\mathbf{x})=g(\mathbf{x}), \ \mathbf{x}\in \mathbf{X}^{\rm te},
$$
where $w(\mathbf{x}, \mathbf{y})$ is a weight function, chosen to be Gaussian: $w(\mathbf{x}, \mathbf{y})=\exp(-\frac{||\mathbf{x}-\mathbf{y}||^2}{\sigma^2})$ with $\sigma$ being a scaling parameter. By taking the variational derivative of the energy functional Eq.~(\ref{DirichletEnergy}), we get the following Euler-Lagrange equation

\begin{equation}
\label{EL-Equation}
\begin{cases}
\sum_{\mathbf{y}\in \mathbf{X}} \left(w(\mathbf{x}, \mathbf{y})+w(\mathbf{y}, \mathbf{x})\right)\left(u(\mathbf{x})-u(\mathbf{y})\right)=0 & \hskip -0.2cm \mathbf{x}\in \mathbf{X}/\mathbf{X}^{\rm te}\\
u(\mathbf{x})=g(\mathbf{x}) & \hskip -0.3cm \mathbf{x}\in \mathbf{X}^{\rm te}.
\end{cases}
\end{equation}
\medskip

\noindent By solving the linear system Eq.~(\ref{EL-Equation}), we obtain labels $u(\mathbf{x})$ for the unlabeled data $\mathbf{x}\in \mathbf{X}/\mathbf{X}^{\rm te}$. The interpolation quality becomes very poor when only a tiny amount of data are labeled, i.e., $|\mathbf{X}^{\rm te}|\ll |\mathbf{X}
/\mathbf{X}^{\rm te}|$. To alleviate this degradation, the weight of the labeled data is increased in the above Euler-Lagrange equation (Eq.~(\ref{EL-Equation})), which gives

\begin{equation}
\label{WNLL}
\begin{cases}
\sum_{\mathbf{y}\in \mathbf{X}} \left(w(\mathbf{x}, \mathbf{y})+w(\mathbf{y}, \mathbf{x})\right)\left(u(\mathbf{x})-u(\mathbf{y})\right)+\\
\left(\frac{|\mathbf{X}|}{|\mathbf{X}^{\rm te}|}-1\right)\sum_{\mathbf{y}\in \mathbf{X}^{\rm te}}w(\mathbf{y}, \mathbf{x})\left(u(\mathbf{x})-u(\mathbf{y})\right)=0 & \hskip -0.2cm \mathbf{x}\in \mathbf{X}/\mathbf{X}^{\rm te}\\
u(\mathbf{x})=g(\mathbf{x}) &\hskip -0.2cm \mathbf{x}\in \mathbf{X}^{\rm te}.
\end{cases}
\end{equation}
\medskip

\noindent We call the solution to Eq.~(\ref{WNLL}) weighted nonlocal Laplacian (WNLL), and denote it as ${\rm WNLL}(\mathbf{X}, \mathbf{X}^{\rm te}, \mathbf{Y}^{\rm te})$. \cite{Shi:2018WNLL}, showed that the WNLL graph interpolating function converges to the solution of the associated high dimensional Laplace-Beltrami equation. For classification, $g(\mathbf{x})$ is the one-hot label for 
$\mathbf{x}$.

\begin{remark}
For a given $\mathbf{x}$, due to the exponential decay of the 
kernel-- $w(\mathbf{x}, \mathbf{y})=\exp(-\frac{||\mathbf{x}-\mathbf{y}||^2}{\sigma^2})$, we do not need to compute weights for all $\mathbf{y}$ in $\mathbf{X}$. In practice, we only consider the contribution from the first $m$-nearest neighbors of $\mathbf{x}$ and let $\sigma$ be the distance between $\mathbf{x}$ and its $n$th nearest neighbor. We use the approximate nearest neighbor \citep{ANN:2014} to search all the $m$ nearest neighbors of any given data $\mathbf{x}$.
\end{remark}

\subsubsection{Theoretical Guarantees for the WNLL Interpolating Function}
To ensure the accuracy of WNLL interpolation, the template data, i.e., the labeled data, should cover all classes of data in $\mathbf{X}$. We give a necessary condition in Theorem \ref{Rep-Thm}.
\begin{theorem}[\cite{BaoWang:2018NIPS}]\label{Rep-Thm}
Suppose we have a dataset, $\mathbf{X}$, which consists of $N$ different classes of data with each instance having the same probability to belong to any of the $N$ classes. Moreover, suppose the number of instances of each class is sufficiently large. If we want to guarantee all classes of data to be sampled at least once, on average at least $N\left(1+\frac{1}{2}+\frac{1}{3}+\cdots +\frac{1}{N}\right)$ data needs to be sampled from $\mathbf{X}$. In this case, the number of data being sampled, in expectation for each class, is $1+\frac{1}{2}+\frac{1}{3}+\cdots +\frac{1}{N} \approx \ln N$.
\end{theorem}
\medskip

\noindent We consider the convergence of the WNLL for graph interpolation and give a theoretical interpretation of the special weight selected in Eq.~(\ref{WNLL}). We summarize some results from \cite{Shi:2018WNLL}. 
Consider the following generalized WNLL interpolation
\begin{equation}
\label{eq:wgl}
\begin{cases}
\sum_{\mathbf{y}\in \mathbf{X}} R_\delta(\mathbf{x},\mathbf{y})\left(u_\delta(\mathbf{x})-u_\delta(\mathbf{y})\right)+ \mu \sum_{\mathbf{y}\in \mathbf{X}^{\rm te}} K_\delta(\mathbf{x}, \mathbf{y})(u_\delta(\mathbf{x})-g(\mathbf{y}))=0,\quad \mathbf{x}\in \mathbf{X},\\
u_\delta(\mathbf{x})=g(\mathbf{x}),\quad \mathbf{x}\in \mathbf{X}^{\rm te}.
\end{cases}
\end{equation}
where $R_\delta(\mathbf{x}, \mathbf{y})$, $K_\delta(\mathbf{x}, \mathbf{y})$ are kernel functions given as
\begin{equation}
\label{eq:kernel}
R_\delta(\mathbf{x}, \mathbf{y}) = C_\delta R\left(\frac{|\mathbf{x} -\mathbf{y}|^2}{4\delta^2}\right), \quad K_\delta(\mathbf{x}, \mathbf{y}) = C_\delta K\left(\frac{|\mathbf{x} - \mathbf{y}|^2}{4\delta^2}\right),
\end{equation}
where $C_\delta = \frac{1}{(4\pi \delta^2)^{k/2}}$ is the normalization factor.
$R, K\in C^2(\mathbb{R}^+) $ are two kernel functions satisfying the conditions listed in Assumption \ref{assumptions}.

\begin{assumption}
\label{assumptions}
\begin{itemize}
\item[]


\item \rm Assumptions on the manifold: $\mathcal{M}$ is a $k$-dimensional closed $C^\infty$ manifold isometrically embedded in a Euclidean space $\mathbb{R}^d$. $\mathcal{D}$ and $\partial\mathcal{D}$ are smooth submanifolds of $\mathbb{R}^d$. Moreover, $g(\mathbf{x})\in C^1(\mathcal{D})$.

\item \rm Assumptions on the kernel functions:
\begin{itemize}
\item[\rm (a)] \rm Smoothness: $K(r), R(r)\in C^2(\mathbb{R}^+)$;
\item[(b)] Nonnegativity: $R(r), K(r)\ge 0$ for any $r\ge 0$.
\item[(c)] Compact support:
$R(r) = 0$ for $\forall r >1$; $K(r) = 0$ for $\forall r > r_0\ge 2$.
\item[(d)] Nondegeneracy:
 $\exists \delta_0>0$ such that $R(r)\ge\delta_0$ for $0\le r\le 1/2$ and $K(r)\ge\delta_0$ for $0\le r\le 2$.
\end{itemize}
\item Assumptions on the point cloud: $\mathbf{X}^{\rm te}$ and $\mathbf{X}$ are uniformly distributed on $\mathcal{M}$ and $\mathcal{D}$, respectively.
\end{itemize}
\end{assumption}
\medskip

\noindent As the continuous counterpart, we consider the Laplace-Beltrami equation on a closed smooth manifold $\mathcal{M}$
\begin{align}
\label{eq:laplace-large}
\left\{
\begin{array}{rcll}
  \Delta_\mathcal{M} u(\mathbf{x})&=&0, & \mathbf{x}\in \mathcal{M},\\
u(\mathbf{x})&=&g(\mathbf{x}), & \mathbf{x}\in \mathcal{D},
\end{array}\right.
\end{align}
where $\Delta_\mathcal{M}=\text{div}(\nabla)$ is the Laplace-Beltrami operator on $\mathcal{M}$. Let $\Phi: \Omega\subset \mathbb{R}^k\rightarrow \mathcal{M}\subset\mathbb{R}^d$ be a local parametrization of $\mathcal{M}$ and $\theta\in \Omega$.
For any differentiable function $f:\mathcal{M}\rightarrow \mathbb{R}$,
we define the gradient on the manifold
\begin{align}
  \label{eq:diff-M}
  \nabla f(\Phi(\theta))&=\sum_{i, j=1}^m g^{ij}(\theta)\frac{\partial \Phi}{\partial\theta_i}(\theta)\frac{\partial f(\Phi(\theta))}{\partial \theta_j}(\theta).
\end{align}
And for the vector field $F:\mathcal{M}\rightarrow T_{\mathbf{x}}\mathcal{M}$ on $\mathcal{M}$, where $T_{\mathbf{x}}\mathcal{M}$ is the tangent space of $\mathcal{M}$ at $\mathbf{x}\in \mathcal{M}$, the divergence is defined as
\begin{align}
\label{eq:diver}
\text{div} (F)&= \frac{1}{\sqrt{\det G}}\sum_{k=1}^d\sum_{i, j=1}^m\frac{\partial}{\partial \theta_i}\left(\sqrt{\det G}\,g^{ij}F^k(\Phi(\theta))\frac{\partial \Phi^k}{\partial\theta_j}\right)
\end{align}
where $(g^{ij})_{i, j=1, \cdots, k}=G^{-1}$, $\det G$ is the determinant of matrix $G$ and $G(\theta)=(g_{ij})_{i,j=1,\cdots,k}$ is the first fundamental form with
\begin{eqnarray}
\label{eq:remainn}
g_{ij}(\theta)=\sum_{k=1}^d\frac{\partial \Phi_k}{\partial \theta_i}(\theta)\frac{\partial \Phi_k}{\partial \theta_j}(\theta),\quad i,j=1,\cdots,m.
\end{eqnarray}
and $(F^1(\mathbf{x}),\cdots,F^d(\mathbf{x}))^T$ is the representation of $F$ in the embedding coordinates.
\medskip

\noindent We have the following high probability guarantee for convergence of the WNLL interpolating function to the solution of the Laplace-Beltrami equation on the manifold $\mathcal{M}$.

\begin{theorem}[\cite{Shi:2018WNLL}]
\label{thm:main}
Let $u_\delta$ solve \eqref{eq:wgl} and $u$ solve \eqref{eq:laplace-large}.
Given Assumption \ref{assumptions}, with probability at least $1-1/(2n),\; where \  n=|\mathbf{X}|$, we have
\begin{equation}
|u_\delta-u|\le C\delta,\nonumber
\end{equation}
as long as
\begin{equation}
\label{eq:condition-mu-intro}
\mu\sum_{\mathbf{y}\in \mathbf{X}^{\rm te}} K_\delta(\mathbf{x}, \mathbf{y})\ge C \sum_{\mathbf{y}\in \mathbf{X}} R_\delta(\mathbf{x}, \mathbf{y}),\quad \mathbf{x}\in \mathbf{X}\cap\mathcal{D}_\delta,
\end{equation}
where $\mathcal{D}_\delta=\{\mathbf{x}\in \mathcal{M}: \text{dist}(\mathbf{x},\mathcal{D})\leq 2\delta\}$, and  $C=C(\mathcal{M}, \mathcal{D}, R, K)>0$ is a constant that is independent of $\delta$, $\mathbf{X}$ and $\mathbf{X}^{\rm te}$.
\end{theorem}

\noindent In the above theorem, Eq.~\eqref{eq:condition-mu-intro} actually gives a constraint on the weight $\mu$. Note that 
$$
\frac{1}{n} \sum_{\mathbf{y}\in \mathbf{X}} R_\delta(\mathbf{x}, \mathbf{y})\approx \frac{1}{|\mathcal{M}| }\int_\mathcal{M} R_\delta(\mathbf{x}, \mathbf{y}) d \mathbf{y}= O(1),\quad \mathbf{x}\in \mathbf{X}\cap\mathcal{D}_\delta.$$
$\mathbf{X}^{\rm te}$ samples $\mathcal{D}$, if $\mathbf{X}^{\rm te}$ is dense enough, we have
\begin{equation}
\frac{1}{|\mathbf{X}^{\rm te}|}\sum_{\mathbf{y}\in \mathbf{X}^{\rm te}} K_\delta(\mathbf{x}, \mathbf{y})\approx\frac{1}{|\mathcal{D}| }\int_\mathcal{D} K_\delta(\mathbf{x}, \mathbf{y}) d \mathbf{y}, \quad \mathbf{x}\in \mathbf{X}\cap\mathcal{D}_\delta.\nonumber
\end{equation}
Here, we need the assumption on $K$ such that $K(r)\geq\delta_0>0, \forall\  0\leq r\leq 2$. This implies that
\begin{equation*}
\int_\mathcal{D} K_\delta(\mathbf{x}, \mathbf{y}) d \mathbf{y} = O(1), \quad \mathbf{x}\in \mathbf{X}\cap\mathcal{D}_\delta.
\end{equation*}
Hence, from Eq.~(\ref{eq:condition-mu-intro}), we have
\begin{equation}
\mu\sim \frac{|\mathbf{X}|}{|\mathbf{X}^{\rm te}|},\nonumber
\end{equation}
which explains the scaling of $\frac{|\mathbf{X}|}{|\mathbf{X}^{\rm te}|}$ in the WNLL.

\subsection{DNNs with the Graph Interpolating Function as Output Activation}
A straightforward approach is to replace the softmax function with the WNLL 
in Fig.~\ref{fig:Standard-DNN-Structure}. However, backpropagation is difficult in this case.
To resolve this, we consider a new DNN architecture as shown in Fig.~\ref{fig:WNLL-DNN-Structure} which will be discussed in detail in Section~\ref{Section:Algorithms}.

\section{Algorithms}\label{Section:Algorithms}
In this section, we will present training and inference algorithms for DNNs with the WNLL as the output activation in both natural and robust fashions. Natural training means to solve the 
ERM problem on the training dataset and robust training stands for training an adversarially robust deep network by solving the EARM problem. Meanwhile, we will also adapt DNNs with the WNLL interpolating output activation to semi-supervised learning.

\subsection{Natural Training and Inference}
We abstract the natural training and testing procedures for DNNs with the WNLL activation in Fig.~\ref{fig:WNLL-DNN-Structure} (a) and (b), respectively. As a prerequisite of the WNLL interpolation, we need to reserve a small portion of data-label pairs, denoted as $(\mathbf{X}^{\rm te}, \mathbf{Y}^{\rm te})$, to interpolate labels for the unlabeled data in both training and testing procedures of DNNs with the WNLL activation. We call $(\mathbf{X}^{\rm te}, \mathbf{Y}^{\rm te})$ as the preserved template. Directly replacing the softmax by the WNLL in the architecture shown in Fig.~\ref{fig:Standard-DNN-Structure} (a) causes difficulties in backpropagation, namely, the gradient $\frac{\partial \mathcal{L}}{\partial \Theta}$ is difficult to compute since WNLL defines a very complex implicit function. Instead, to train DNNs with the WNLL as the output activation, we propose a proxy via an auxiliary neural net (Fig. \ref{fig:WNLL-DNN-Structure} (a)). On top of the original DNNs, we add a buffer block (a fully connected layer followed by a ReLU) and followed by two parallel branches, the WNLL and the linear (fully connected) layers. We train the auxiliary DNNs by alternating between the following two steps: training DNNs with linear and WNLL activation, respectively. In the following, we denote DNN with the WNLL activation as DNN-WNLL, e.g., we denote ResNet20 with WNLL activation as ResNet20-WNLL.
\medskip

\noindent {\bf Train DNN-WNLL with linear activation: } Run $N_1$ steps of the following forward and backward propagation, where in the $k$th iteration, we have:
\begin{itemize}
\item {\it Forward propagation:} Transform the training data $\mathbf{X}$, respectively, by DNN, Buffer and Linear blocks into the predicted labels $\tilde{\mathbf{Y}}$:
$$\tilde{\mathbf{Y}} = {\rm Linear}({\rm Buffer}({\rm DNN}(\mathbf{X}, \Theta^{k-1}), \mathbf{W}_B^{k-1}), \mathbf{W}_L^{k-1}).$$
Then compute the loss between the ground truth labels $\mathbf{Y}$ and the predicted ones $\tilde{\mathbf{Y}}$, denoted the loss as $\mathcal{L}^{\rm Linear}$.

\item {\it Backpropagation:} Update 
($\Theta^{k-1}$, $\mathbf{W}_B^{k-1}$, $\mathbf{W}_L^{k-1}$) by stochastic gradient descent:
$$
\mathbf{W}_L^k = \mathbf{W}_L^{k-1} - \gamma \frac{\partial \mathcal{L}^{\rm Linear}}{\partial \tilde{\mathbf{Y}}}\cdot \frac{\partial \tilde{\mathbf{Y}}}{\partial \mathbf{W}_L},
$$
$$
\mathbf{W}_B^k = \mathbf{W}_B^{k-1} - \gamma \frac{\partial \mathcal{L}^{\rm Linear}}{\partial \tilde{\mathbf{Y}}}\cdot \frac{\partial \tilde{\mathbf{Y}}}{\partial \hat{\mathbf{X}}}\cdot \frac{\partial \hat{\mathbf{X}}}{\partial \mathbf{W}_B},
$$
$$
\Theta^k = \Theta^{k-1} - \gamma \frac{\partial \mathcal{L}^{\rm Linear}}{\partial \tilde{\mathbf{Y}}}\cdot \frac{\partial \tilde{\mathbf{Y}}}{\partial \hat{\mathbf{X}}}\cdot \frac{\partial \hat{\mathbf{X}}}{\partial \tilde{\mathbf{X}}}\cdot \frac{\partial \tilde{\mathbf{X}}}{\partial \Theta}.
$$
\end{itemize}

\noindent {\bf Train DNN-WNLL with the WNLL activation:} Run $N_2$ steps of the following forward and backward propagation, where in the $k$th iteration, we have:
\begin{itemize}
\item {\it Forward propagation:} The training data $\mathbf{X}$, template $\mathbf{X}^{\rm te}$ and $\mathbf{Y}^{\rm te}$ are transformed, respectively, by DNN, Buffer, and WNLL blocks to get predicted labels $\hat{\mathbf{Y}}$:
$$\hat{\mathbf{Y}} = {\rm WNLL}({\rm Buffer}({\rm DNN}(\mathbf{X}, \Theta^{k-1}), \mathbf{W}_B^{k-1}), \hat{\mathbf{X}}^{\rm te}, \mathbf{Y}^{\rm te}).$$
Then compute the loss, $\mathcal{L}^{\rm WNLL}$, between the ground truth labels $\mathbf{Y}$ and predicted ones $\hat{\mathbf{Y}}$.

\item {\it Backpropagation:} Update weights $\mathbf{W}_B^{k-1}$ only, $\mathbf{W}_L^{k-1}$ and $\Theta^{k-1}$ will be tuned in the next iteration in training DNN-WNLL with the linear activation, by stochastic gradient descent.

\begin{eqnarray}
\mathbf{W}_B^{k} &=&  \mathbf{W}_B^{k-1} - \gamma \frac{\partial \mathcal{L}^{\rm WNLL}}{\partial \hat{\mathbf{Y}}}\cdot \frac{\partial \hat{\mathbf{Y}}}{\partial \hat{\mathbf{X}}}\cdot \frac{\partial \hat{\mathbf{X}}}{\partial \mathbf{W}_B} \\ \nonumber
&\approx& \mathbf{W}_B^{k-1} - \gamma \frac{\partial \mathcal{L}^{\rm Linear}}{\partial \tilde{\mathbf{Y}}}\cdot \frac{\partial \tilde{\mathbf{Y}}}{\partial \hat{\mathbf{X}}}\cdot \frac{\partial \hat{\mathbf{X}}}{\partial \mathbf{W}_B}.
\label{eq:bp-wnll}
\end{eqnarray}
\end{itemize}
\medskip

\noindent We use the computational graph of the left branch (linear layer) to compute the approximated gradients for the DNN with WNLL activation. For a given loss value $\mathcal{L}^{\rm WNLL}$, we adopt the approximation
$\frac{\partial \mathcal{L}^{\rm WNLL}}{\partial \hat{\mathbf{Y}}}\cdot \frac{\partial \hat{\mathbf{Y}}}{\partial \hat{\mathbf{X}}} \approx \frac{\partial \mathcal{L}^{\rm Linear}}{\partial \tilde{\mathbf{Y}}}\cdot \frac{\partial \tilde{\mathbf{Y}}}{\partial \hat{\mathbf{X}}}$ where the right hand side is also evaluated at the value of $\mathcal{L}^{\rm WNLL}$.
The heuristic behind this approximation is the following: WNLL defines a harmonic function implicitly, and the linear function is the simplest nontrivial explicit harmonic function. Empirically, we observe this simple approximation works well in training the deep network. The reason why we freeze the network in the DNN block is mainly because of the stability concerns.
\medskip

\noindent The above alternating scheme is an algorithm of a greedy fashion. During training, the WNLL activation plays two roles: on the one hand, alternating between the linear and the WNLL activation benefits both which enables the neural nets to learn features that are appropriate for both linear classification and the WNLL based manifold interpolation. On the other hand, in the case when we lack sufficient training data, the training of DNNs usually gets stuck at some bad local minima which cannot generalize well on the new data. We use the WNLL interpolation to perturb those learned sub-optimal weights and to help to arrive at a local minimum with better generalizability. At inference (test) time, we remove the linear classifier from the neural nets and use the DNN block together with the WNLL to predict new data (Fig.~\ref{fig:WNLL-DNN-Structure} (b)). The reason for using the WNLL instead of the linear layer is simply because the WNLL interpolation is superior to the linear classifier and this superiority is preserved when applied to deep features (which will be confirmed in Section.~\ref{Section:Results}). Moreover, the WNLL interpolation utilizes both deep learning features and the reserved template at the test time to guide the classifier and to enhance adversarial robustness in classification.
\medskip

\noindent We summarize the training and testing for DNN-WNLL in Algorithms~\ref{alg-Train} and \ref{alg-Test}, respectively. In each round of the alternating procedure, i.e., each outer loop in Algorithm~\ref{alg-Train}, the entire training dataset $(\mathbf{X}, \mathbf{Y})$ is first used to train DNN-WNLL with the linear activation. We randomly separate a template, e.g., half of the entire data from the training set which will be used to perform WNLL interpolation in training DNN-WNLL with the WNLL activation. In practice, for both training and testing, we use mini-batches for both the template and the interpolated points when the entire dataset is too large. The final predicted labels are based on a majority voting across interpolation results from all the template mini-batches. 

\begin{algorithm}
\caption{DNN with the WNLL Output Activation: Training Procedure.}\label{alg-Train}
\begin{algorithmic}[1]
\State \textbf{Input:} Training set: (data, label) pairs $(\mathbf{X}, \mathbf{Y})$. The number of alternating steps $N$ and the number of epochs for training DNN with WNLL activation $M$.
\State \textbf{Output:} An optimized DNN with the WNLL activation, denoted as DNN-WNLL. 
\For {${\rm iter} = 1, $\dots$, N$ 
}
\State //Train the left branch: DNN with the linear activation.
\State Train DNN $+$ Linear blocks, and denote the learned model as
DNN-Linear. 
\State //Train the right branch: DNN with the WNLL activation.
\State Split $(\mathbf{X}, \mathbf{Y})$ into training data and template, i.e., $(\mathbf{X}, \mathbf{Y}) \doteq (\mathbf{X}^{\rm tr}, \mathbf{Y}^{\rm tr}) \bigcup (\mathbf{X}^{\rm te}, \mathbf{Y}^{\rm te})$.
\State Partition the training data into $M$ mini-batches, i.e., $(\mathbf{X}^{\rm tr}, \mathbf{Y}^{\rm tr}) = \bigcup_{i=1}^M (\mathbf{X}_i^{\rm tr}, \mathbf{Y}_i^{\rm tr})$.
\For {$i = 1, 2, \cdots, M$}
\State Transform $\mathbf{X}_i^{\rm tr}\bigcup \mathbf{X}^{\rm te}$ by DNN-Linear, 
i.e.,  $\tilde{\mathbf{X}}^{\rm tr} \bigcup \tilde{\mathbf{X}}^{\rm te} = {\rm DNN}_{\rm Linear}(\mathbf{X}_i^{\rm tr}\bigcup \mathbf{X}^{\rm te})$.
\State Apply WNLL (Eq.(\ref{WNLL})) on $\{\tilde{\mathbf{X}}^{\rm tr} \bigcup \tilde{\mathbf{X}}^{\rm te}, \mathbf{Y}^{\rm te}\}$ to interpolate label $\tilde{\mathbf{Y}}^{\rm tr}$.
\State Backpropagate the error between $\mathbf{Y}^{\rm tr}$ and $\tilde{\mathbf{Y}}^{\rm tr}$ via Eq.(\ref{eq:bp-wnll}) to update $\mathbf{W}_B$ only.
\EndFor
\EndFor
\end{algorithmic}
\end{algorithm}


\begin{remark}
In Algorithm~\ref{alg-Train}, the WNLL interpolation is also performed in a mini-batch manner (as shown in the inner iteration). Based on our experiments, this has a very small influence on reducing interpolation accuracy.
\end{remark}

\begin{algorithm}
\caption{DNN with the WNLL Output Activation: Testing Procedure.}\label{alg-Test}
\begin{algorithmic}[1]
\State \textbf{Input:} Testing data $\mathbf{X}$, template $(\mathbf{X}^{\rm te}, \mathbf{Y}^{\rm te})$. The optimized model 
DNN-WNLL. 
\State \textbf{Output:} Predicted label $\tilde{\mathbf{Y}}$ for $\mathbf{X}$.
\State Apply the DNN block of the DNN-WNLL 
to $\mathbf{X}\bigcup \mathbf{X}^{\rm te}$ to get the features $\tilde{\mathbf{X}}\bigcup \tilde{\mathbf{X}}^{\rm te}$.
\State Apply the WNLL interpolation (Eq.(\ref{WNLL})) on $\{\tilde{\mathbf{X}} \bigcup \tilde{\mathbf{X}}^{\rm te}, \mathbf{Y}^{\rm te}\}$ to interpolate label $\tilde{\mathbf{Y}}$.
\end{algorithmic}
\end{algorithm}

\subsection{Adversarial Training}
Adversarial training is one of the most generic frameworks for adversarial defense. The key idea of adversarial training is to augment the training data with adversarial versions which can be obtained by applying adversarial attacks to the clean data. In the following, we adopt the minimax formalism of the adversarial training proposed by \cite{Madry:2018}.

\subsubsection{Adversarial Attacks}
We consider three benchmark attacks: the fast gradient sign method (FGSM) and the iterative fast gradient sign method (IFGSM) in the $\ell_\infty$-norm \citep{Goodfellow:2014AdversarialTraining}, and the \cite{CWAttack:2016} attack in the $\ell_2$-norm (C\&W). We denote the classifier defined by a specific DNN as $\tilde{y} = f(\Theta, \mathbf{x})$ for a given instance ($\mathbf{x}$, $y$). FGSM searches the adversarial image $\mathbf{x}'$ by maximizing the loss $\mathcal{L}(\mathbf{x}', y) \doteq \mathcal{L}(f(\Theta, \mathbf{x}'), y)$ with a maximum allowed perturbation $\epsilon$, i.e., $\|\mathbf{x}'-\mathbf{x}\|_\infty \leq \epsilon$. We can approximately solve this constrained optimization problem by linearize the objective function, i.e.,
$$
\mathcal{L}(\mathbf{x}', y) \approx \mathcal{L}(\mathbf{x}, y) + \nabla_\mathbf{x}\mathcal{L}(\mathbf{x}, y)^T \cdot (\mathbf{x}'-\mathbf{x}).
$$ 
Under this linear approximation, the optimal adversarial image is
\begin{equation}
\label{FGSM}
\mathbf{x}'=\mathbf{x} + \epsilon \, {\rm sign} \left( \nabla_\mathbf{x}\mathcal{L}(\mathbf{x}, y) \right).
\end{equation}
\medskip

\noindent IFGSM iterates FGSM to generate the enhanced adversarial images, where the iteration proceeds as follows
\begin{equation}
\label{IFGSM-1}
\mathbf{x}^{(m)} = \mathbf{x}^{(m-1)} + \alpha \cdot {\rm sign} \left( \nabla_{\mathbf{x}} \mathcal{L}(\mathbf{x}^{(m-1)}, y) \right),
\end{equation}
where $m=1, \cdots, M$, $\mathbf{x}^{(0)}=\mathbf{x}$ and $\alpha$ being the step size. Moreover, let the adversarial image be $\mathbf{x}'=\mathbf{x}^{(M)}$ with $M$ being the number of iterations. To ensure the maximum perturbation to the clean image is no bigger than $\epsilon$, in each iteration we clip the intermediate adversarial images which results in the following attack scheme
\begin{equation}
\label{IFGSM}
\mathbf{x}^{(m)} = {\rm Clip}_{\mathbf{x}, \epsilon}\left\{\mathbf{x}^{(m-1)} + \alpha \cdot {\rm sign} \left( \nabla_{\mathbf{x}} \mathcal{L}(\mathbf{x}^{(m-1)}, y) \right)\right\},
\end{equation}
where ${\rm Clip}_{\mathbf{x}, \epsilon}(\mathbf{x}')$ limits the change of the generated adversarial image in each iteration, and it is defined as
$$
{\rm Clip}_{\mathbf{x}, \epsilon}(\mathbf{x}') = \min\left\{1, \mathbf{x}+\epsilon, \max\{0, \mathbf{x}-\epsilon, \mathbf{x}'\} \right\},
$$
where we assume the pixel value of the image is normalized to $[0, 1]$.
\medskip

\noindent Both FGSM and IFGSM belong to the fixed-perturbation attacks. Moreover, we consider a zero-confidence attack proposed by Carlini and Wagner. For a given image-label pair $(\mathbf{x}, y)$, and for any given label $t\neq y$, C\&W attack searches the adversarial image that will be classified to class $t$ with minimum perturbation by solving the following optimization problem
\begin{equation}
\label{cwl2-eq1}
\min_{\boldsymbol{\delta}} ||\boldsymbol{\delta}||_2^2,
\end{equation}
subject to
$$
f(\mathbf{x}+\boldsymbol{\delta}) = t, \; \mathbf{x}+\boldsymbol{\delta} \in [0, 1]^n,
$$
where $\boldsymbol{\delta}$ is the adversarial perturbation (for the sake of simplicity, we ignore the dependence on $\Theta$ in $f$). The equality constraint in Eq.~(\ref{cwl2-eq1}) is hard to tackle, so Carlini and Wagner considered the following surrogate constraint
\begin{equation}
\label{cwl2-eq2}
g(\mathbf{x}) = \max\left(\max_{i\neq t}(Z(\mathbf{x})_i) - Z(\mathbf{x})_t , 0\right),
\end{equation}
where $Z(\mathbf{x})$ is the logit vector for an input $\mathbf{x}$, i.e., output of the neural net before the output layer, and $Z(\mathbf{x})_i$ is the logit value corresponding to class $i$. It is easy to see that $f(\mathbf{x}+\boldsymbol{\delta})=t$ is equivalent to $g(\mathbf{x}+\boldsymbol{\delta})\leq 0$. Therefore, the problem in Eq.~(\ref{cwl2-eq1}) can be reformulated as
\begin{equation}
\label{cwl2-eq4}
\min_{\boldsymbol{\delta}} ||\boldsymbol{\delta}||_2^2 + c \cdot g(\mathbf{x}+\boldsymbol{\delta}),
\end{equation}
subject to
$$
\mathbf{x}+\boldsymbol{\delta} \in [0, 1]^d,
$$
where $c\geq 0$ is the Lagrangian multiplier. 
\medskip

\noindent By letting $\boldsymbol{\delta} = \frac{1}{2}\left(\tanh(\mathbf{w})+1\right)-\mathbf{x}$, Eq.~(\ref{cwl2-eq4}) can be written as an unconstrained optimization problem. Moreover, Carlini and Wagner introduced the confidence parameter $\kappa$ into the above formulation. In a nutshell, the C\&W attack seeks the adversarial image by solving the following problem
\begin{align}
\label{CWL2}
&\min_{\mathbf{w}} \|\frac{1}{2}\left(\tanh(\mathbf{w}) + 1\right) - \mathbf{x} \|_2^2 + c\cdot\\ \nonumber
&\max\left\{-\kappa, \max_{i\neq t}(Z(\frac{1}{2}(\tanh(\mathbf{w}))+1)_i) 
- Z(\frac{1}{2}(\tanh(\mathbf{w}))+1)_t \right\}.
\end{align}
The Adam optimizer \citep{Kingma:2014Adam} can solve the above unconstrained optimization problem, Eq.~(\ref{CWL2}), efficiently. All three attacks clip the values of each pixel of the adversarial image $\mathbf{x}'$ to between 0 and 1. 
\medskip

\noindent The only difficulty in extending the above three adversarial attacks to DNN-WNLL is again to compute the gradient in backpropagation. Similar to the training of DNN-WNLL, we compute the following surrogate gradient by linearizing the WNLL activation. For a given mini-batch of test image-label pairs $(\mathbf{X}, \mathbf{Y})$ and template $(\mathbf{X}^{\rm te}, \mathbf{Y}^{\rm te})$, we denote the DNN-WNLL as $\tilde{\mathbf{Y}} = {\rm WNLL}(Z(\{\mathbf{X}, \mathbf{X}^{\rm te}\} ), \mathbf{Y}^{\rm te})$, where $Z(\{\mathbf{X}, \mathbf{X}^{\rm te}\} )$ is the composition of the DNN and buffer blocks as shown in Fig.~\ref{fig:WNLL-DNN-Structure} (a). By ignoring dependence of the loss function on the parameters, the loss function for DNN-WNLL can be written as $\tilde{\mathcal{L}}(\mathbf{X}, \mathbf{Y}, \mathbf{X}^{\rm te}, \mathbf{Y}^{\rm te}) \doteq {\rm Loss}(\hat{\mathbf{Y}}, \mathbf{Y})$. The above three attacks for DNN-WNLL are summarized below.

\begin{itemize}
    \item {\bf FGSM} 
    
    \begin{align}
    \label{FGSM-WNLL}
    \mathbf{X}' = \mathbf{X} + \epsilon \cdot {\rm sign}\left( \nabla_\mathbf{X} \tilde{\mathcal{L}}(\mathbf{X}, \mathbf{Y}, \mathbf{X}^{\rm te}, \mathbf{Y}^{\rm te}) \right).
    \end{align}
    
    \item {\bf IFGSM} 
    \begin{align}
    \label{IFGSM-WNLL}
    \mathbf{X}^{(m)} = {\rm Clip}_{\mathbf{X}, \epsilon}[\mathbf{X}^{(m-1)} +
    \alpha \cdot {\rm sign} \left( \nabla_{\mathbf{X}} \tilde{\mathcal{L}}(\mathbf{X}^{(m-1)}, \mathbf{Y}, \mathbf{X}^{\rm te}, \mathbf{Y}^{\rm te}) \right)],
    \end{align}
    where $m=1, 2, \cdots, M$;  $\mathbf{X}^{(0)} = \mathbf{X}$ and $\mathbf{X}'=\mathbf{X}^{(M)}$.
    
    \item {\bf C\&W}
    \begin{align}\label{CWL2-WNLL}
    \min_{\mathbf{W}}  ||\frac{1}{2}\left(\tanh(\mathbf{W}) + 1\right) - \mathbf{X} ||_2^2  + \\ \nonumber
    c\cdot  \max[-\kappa,
       \max_{\mathbf{i}\neq \mathbf{t} }( Z(\frac{1}{2}(\tanh(\mathbf{W}))+1)_\mathbf{i}) -  Z(\frac{1}{2}(\tanh(\mathbf{W}))+1)_\mathbf{t} ],
    \end{align}
    where $\mathbf{i}$ are the logit values of the input images $\mathbf{X}$, $\mathbf{t}$ are the target labels. 
\end{itemize}
\medskip

\noindent In the above attacks, $\nabla_{\mathbf{X}}\tilde{\mathcal{L}}$ is required to generate the adversarial images. In the DNN-WNLL, this gradient is difficult to compute. As shown in Fig.~\ref{fig:WNLL-DNN-Structure} (b), we approximate $\nabla_{\mathbf{X}}\tilde{\mathcal{L}}$ in the following way
\begin{eqnarray}
\label{GradientApprox}
\nabla_{\mathbf{X}}\tilde{\mathcal{L}} = \frac{\partial \mathcal{L}^{\rm WNLL}}{\partial\hat{\mathbf{Y}}}\cdot \frac{\partial\hat{\mathbf{Y}}}{\partial\hat{\mathbf{X}}} \cdot \frac{\partial\hat{\mathbf{X}}}{\partial\tilde{\mathbf{X}}} \cdot \frac{\partial\tilde{\mathbf{X}}}{\partial\mathbf{X}} \\ \nonumber
\approx
\frac{\partial \mathcal{L}^{\rm Linear}}{\partial\tilde{\mathbf{Y}}}\cdot \frac{\partial\tilde{\mathbf{Y}}}{\partial\hat{\mathbf{X}}} \cdot \frac{\partial\hat{\mathbf{X}}}{\partial\tilde{\mathbf{X}}} \cdot \frac{\partial\tilde{\mathbf{X}}}{\partial\mathbf{X}},
\end{eqnarray}
again, in the above approximation, we set the value of $\mathcal{L}^{\rm Linear}$ to that of $\tilde{\mathcal{L}}$.
\medskip

\noindent Based on our numerical experiments, the batch size of $\mathbf{X}$ has a negligible influence on the adversarial attack and defense. In all of our experiments, we choose the size of both mini-batches $\mathbf{X}$ and the template to be $500$.

\subsubsection{Adversarial Training}
We apply the projected gradient descent (PGD) adversarial training \citep{Madry:2018} to train the adversarially robust DNNs, where we approximately solve the EARM (Eq.~(\ref{Adversarial-ERM}) by using the PGD adversarial images, i.e., IFGSM attacks with an initial random perturbation on the clean images, to approximate the solution of the inner maximization problem. We summarize the PGD adversarial training 
for DNNs with the WNLL activation, as shown in Fig.~\ref{fig:WNLL-DNN-Structure} (a), in Algorithm~\ref{PGD-WNLL}.

\begin{algorithm*}[t]
\caption{DNN with the WNLL Output Activation: PGD Adversarial Training}\label{PGD-WNLL}
\begin{algorithmic}[1]
\State \textbf{Input:} Training set: (data, label) pairs $(\mathbf{X}, \mathbf{Y})$, the number of PGD iterations $M$, PGD attack step size $\alpha$, maximum PGD perturbation $\epsilon$. The number of alternating iterations $N$, and the number of epochs used to train DNN with the linear activation $N_1$ and the WNLL activation $N_2$.
\State \textbf{Output:} An optimized DNN-WNLL.
\For {${\rm iter} = 1, \dots, N$ 
}
\State //PGD adversarial training of the left branch: DNN with linear activation.
\State Train DNN $+$ Linear blocks.
\State Partition the training data into $M_1$ mini-batches, i.e., $(\mathbf{X}, \mathbf{Y}) = \bigcup_{i=1}^{M_1} (\mathbf{X}_i, \mathbf{Y}_i)$.
\For{${\rm epoch_1} = 1, $\dots$, N_1$ 
}
\For{$i = 1, $\dots$, M_1$}
\State //{Attack the input images by PGD attack.}
\State $\mathbf{X}_i = \mathbf{X}_i + \mathbf{U}(-\epsilon, \epsilon)$ with $\mathbf{U}(-\epsilon, \epsilon)$ be a uniform random vector.
\For{${\rm iter_1} = 1, \dots,  M$}
\State Attack $\mathbf{X}_i$ according to Eq.~(\ref{IFGSM}).
\EndFor
\State Backpropagate the classification error of the adversarial images.
\EndFor
\EndFor

\State //{PGD adversarial training of the right branch: DNN with WNLL activation.}
\State Split $(\mathbf{X}, \mathbf{Y})$ into training data and template, i.e., $(\mathbf{X}, \mathbf{Y}) \doteq (\mathbf{X}^{\rm tr}, \mathbf{Y}^{\rm tr}) \bigcup (\mathbf{X}^{\rm te}, \mathbf{Y}^{\rm te})$.
\State Partition the training data into $M_2$ mini-batches, i.e., $(\mathbf{X}^{\rm tr}, \mathbf{Y}^{\rm tr}) = \bigcup_{i=1}^{M_2} (\mathbf{X}_i^{\rm tr}, \mathbf{Y}_i^{\rm tr})$.
\For{${\rm epoch_2} = 1, $\dots$, N_2$ 
}
\For{$i=1, \dots, M_2$}
\State //{Attack the input training images by PGD attack.}
\State $\mathbf{X}_i^{\rm tr} = \mathbf{X}_i^{\rm tr} + \mathbf{U}(-\epsilon, \epsilon)$.
\For{${\rm iter_1} = 1, \dots,  M$ 
}
\State Attack $\mathbf{X}^{\rm tr}_i$ according to Eq.~(\ref{IFGSM-WNLL}).
\EndFor
\State Backpropagate the classification error of the adversarial images.
\EndFor
\EndFor
\EndFor
\end{algorithmic}
\end{algorithm*}

\subsection{Semi-supervised Learning}
Semi-supervised learning is another fundamental learning paradigm, where we have access to a large amount of training data. However, most of the training data is unlabeled. Semi-supervised learning is of particular importance in e.g., medical applications \citep{Olivier:2006}.
It is straightforward to extend DNNs with the WNLL activation to semi-supervised learning. Let the labeled and unlabeled training data be $\{\mathbf{X}^{l}, \mathbf{Y}^{l}\}$ and $\{\mathbf{X}^{ul}, \mathbf{Y}^{ul}\}$, respectively. There are two approaches to semi-supervised learning by using DNN-WNLL.
\begin{itemize}
    \item {\bf Approach I:} Train DNN-WNLL on only labeled data $\{\mathbf{X}^{l}, \mathbf{Y}^{l}\}$. During testing, we feed the unlabeled data together with the labeled template data to predict labels for the testing data. This is essentially similar to the classical graph Laplacian-based semi-supervised learning on the deep learning features.
    
    \item {\bf Approach II:} Train DNNs with the WNLL activation by using both labeled $\{\mathbf{X}^{l}, \mathbf{Y}^{l}\}$ and unlabeled $\{\mathbf{X}^{ul}, \mathbf{Y}^{ul}\}$ data. During training, we use both labeled and unlabeled data to build a graph for WNLL interpolation, and then we backpropagate loss between predicted and true labels of the labeled data. The testing phase is the same as that in {\bf Approach I}.
\end{itemize}

\noindent In this work, we focus on the {\bf Approach I}.

\section{Numerical Results}\label{Section:Results}
In this section, we will numerically verify the accuracy and robustness of DNN-WNLL. Moreover, we show that DNN-WNLL is suitable for data-efficient learning. 
We also provide results of semi-supervised learning by using DNN-WNLL. We implement our algorithm on the PyTorch platform \citep{paszke2017automatic}. All the computations are carried out on a machine with a single Nvidia Titan Xp graphics card.
\medskip

\noindent To validate the classification accuracy, efficiency, and robustness of the proposed framework, we test the new architecture and algorithm on the CIFAR10, CIFAR100 \citep{Cifar:2009}, MNIST \citep{MNIST:1998} and SVHN datasets \citep{SVHN:2011}. In all the experiments below, we apply the standard data augmentation that is used for the CIFAR datasets \citep{ResNet,DenseNet,Zagoruyko2016WRN}. For MNIST and SVHN, we use the raw data without any data augmentation. 
\medskip

\noindent Before diving into the performance of DNNs with different output activation functions, we first compare the performance of the WNLL with the softmax on the raw input images for various datasets. The training sets are used to train the softmax classifier and interpolate labels for the test set in the WNLL interpolation, respectively. Table~\ref{Simple-Classifiers} lists the classification accuracies of the WNLL and the softmax on three datasets. For the WNLL interpolation, we only use the top $30$ nearest neighbors to ensure sparsity of the weight matrix to speed up the computation, and the $15$th neighbor's distance is used to normalize the weight matrix. WNLL outperforms softmax remarkably in all the three benchmark tasks especially for the MNIST (Test accuracy: $92.65$\% v.s. $97.74$\%) and SVHN (Test accuracy: $24.66$\% v.s. $56.17$\%) classification. These results indicate potential benefits of using the WNLL instead of the softmax as the output activation in DNNs.
\medskip

\begin{table}[!ht]
\centering
\fontsize{8.5}{8.5}\selectfont
\begin{threeparttable}
\caption{Accuracies of the softmax and the WNLL classifiers in classifying some benchmark datasets.}
\label{Simple-Classifiers}
\begin{tabular}{cccc}
\toprule[1.0pt]
\ \ \ \ \ \ \ \ Dataset\ \ \ \ \ \ \ \  &\ \ \ \ \ \ \ \  CIFAR10\ \ \ \ \ \ \ \  &\ \ \ \ \ \ \ \  MNIST\ \ \ \ \ \ \ \  &\ \ \ \ \ \ \ \  SVHN\ \ \ \ \ \ \ \ \cr
\midrule[0.8pt]
softmax & 39.91\%  & 92.65\%   & 24.66\% \cr
WNLL    & 40.73\%  & 97.74\%   & 56.17\% \cr
\bottomrule[1.0pt]
\end{tabular}
\end{threeparttable}
\end{table}

\noindent For natural training of the DNN-WNLL: We take two passes of the alternating step, i.e., set $N=2$ in Algorithm \ref{alg-Train}. For training of the linear activation stage (Stage 1), we train the network for $n=400$ epochs with stochastic gradient descent. For the training of the WNLL activation stage (Stage 2) we train for $n=5$ epochs. In the first pass, the initial learning rate is $0.05$ and halved after every $50$ epoch in training DNNs with linear activation, and a fixed learning rate $0.0005$ is used to train DNNs with the WNLL activation. The same Nesterov momentum and weight decay as that used in \citep{ResNet,Huang:2016ECCV} are employed for the CIFAR and the SVHN experiments, respectively, in our work. In the second pass, the learning rate is set to be one-fifth of the corresponding epochs in the first pass. The batch sizes are $128$ and $2000$ when training softmax/linear and WNLL activated DNNs, respectively. For a fair comparison, we train the vanilla DNNs with the softmax output activation for $810$ epochs with the same optimizer used in the WNLL activated ones. 


\subsection{Data Efficient Learning -- Small Training Data Case}
\begin{figure}[!ht]
\centering
\begin{tabular}{cc}
\includegraphics[width=0.46\columnwidth]{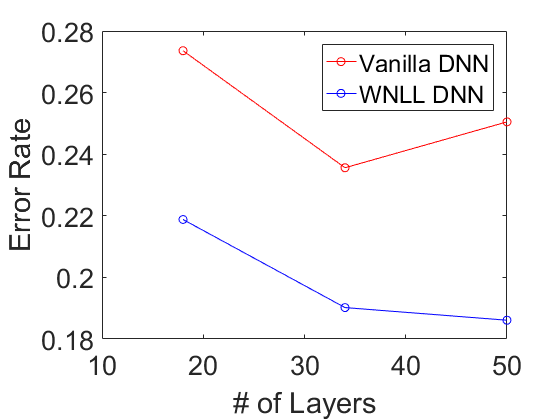}&
\includegraphics[width=0.46\columnwidth]{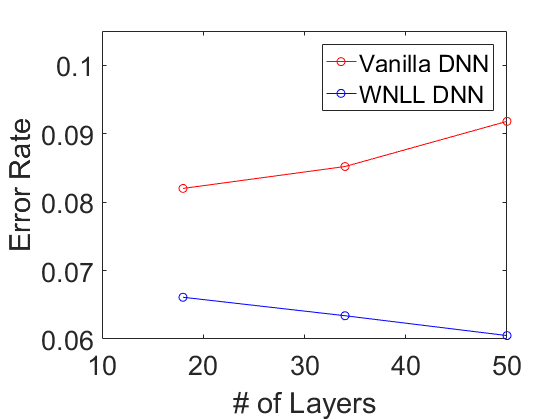}\\
(a)&(b)\\
\end{tabular}
\caption{Plots of test errors when $1$K (a) and $10$K (b) training data are used to train the vanilla and the WNLL activated DNNs. In each plot, we test three different deep networks: PreActResNet18, PreActResNet34, and PreActResNet50. All tests are done on the CIFAR10 dataset.}
\label{Degenerate}
\end{figure}

When we do not have a sufficient amount of labeled training data to train a high capacity deep network, the generalization accuracy of the trained model typically decays as the network goes deeper. We illustrate this in Fig.~\ref{Degenerate}. The WNLL activated DNNs, with its superior regularization power and perturbation capability on bad local minima, can overcome this generalization degradation. 
The left and right panels of Fig.~\ref{Degenerate} plot the results of DNNs with the softmax and the WNLL activation that are trained on $1$K and $10$K images, respectively. These results show that the generalization error rate decays consistently as the network goes deeper in DNN-WNLL. Moreover, the generalization accuracy between the vanilla and the WNLL activated DNNs can differ up to $10$ percent within our testing regime.
\medskip

\begin{figure}[!ht]
\centering
\begin{tabular}{cc}
\includegraphics[width=0.46\columnwidth]{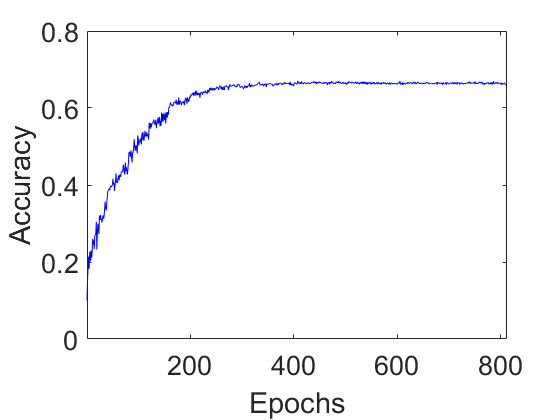}&
\includegraphics[width=0.46\columnwidth]{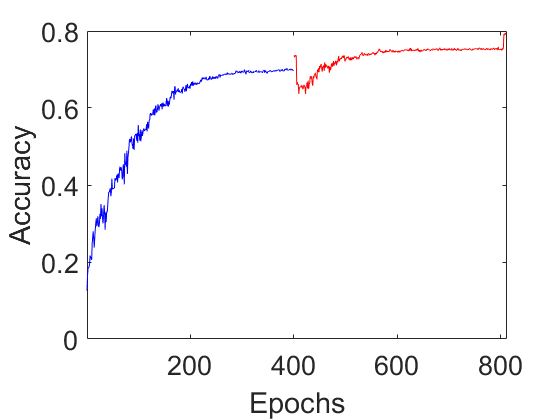}\\
(a)&(b)\\
\includegraphics[width=0.46\columnwidth]{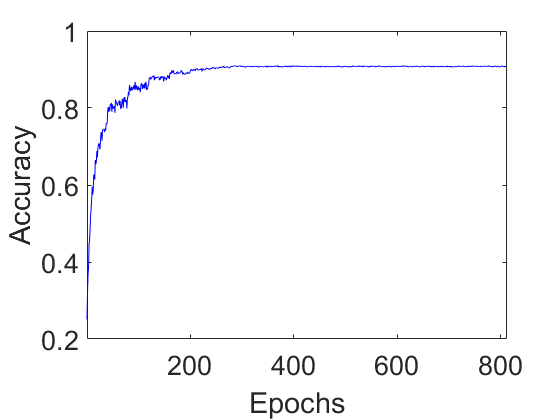}&
\includegraphics[width=0.46\columnwidth]{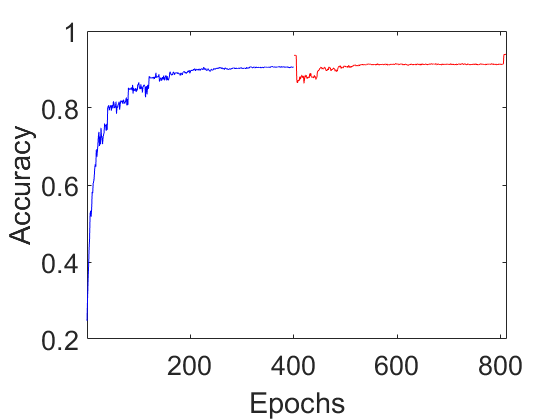}\\
(c)&(d)\\
\end{tabular}
\caption{Evolution of the generation accuracy over the training procedure. Charts (a) and (b) plot the accuracy evolution of ResNet50 with the softmax and the WNLL activation trained with $1$K training data, respectively. Panels (c) and (d) correspond to the case of 10K training data for PreActResNet50. All tests are done on the CIFAR10 dataset.}
\label{Generation-Acc-Evolution}
\end{figure}

\noindent Figure~\ref{Generation-Acc-Evolution} plots the evolution of generalization accuracy during training. We compute the test accuracy per epoch. Panels (a) and (b) plot the test accuracies for the ResNet50 with the softmax and the WNLL activation (1-400 and 406-805 epochs corresponds to linear activation), respectively, with only the first $1$K instances in the training set of CIFAR10, are used to train the models. Charts (c) and (d) are the corresponding plots with $10$K training instances, using a pre-activated ResNet50. After around $300$ epochs, the accuracies of the vanilla DNNs plateau and cannot improve anymore. However, the test accuracy for WNLL jumps at the beginning of Stage 2 in the first pass; during the Stage 1 of the second pass, even though initially there is an accuracy reduction, the accuracy continues to climb and eventually surpasses that of the WNLL activation in Stage 2 of the first pass. The jumps in accuracy at epoch 400 and 800 are due to switching from linear activation to WNLL for predictions on the test set. The initial decay when alternating back to the softmax is caused partially by the final layer $W_L$ not being tuned with respect to the deep features $\tilde{\mathbf{X}}$, and partially due to predictions on the test set being made by the softmax instead of the WNLL. Nevertheless, the perturbation via the WNLL activation quickly results in the accuracy increasing beyond the linear stage in the previous pass.

\subsection{Generalization of Naturally Trained DNN-WNLL}
We next show the superiority of the DNN-WNLL in terms of generalization accuracies when compared to their surrogates with the softmax or the SVM output activation functions. Besides ResNets, we also test the WNLL surrogate on the VGG networks. In table~\ref{Cifar10}, we list the generalization errors for $15$ different DNNs from VGG, ResNet, Pre-activated ResNet families trained on the entire, first $10$K and first $1$K instances of the CIFAR10 training set. We observe that WNLL, in general, improves more for ResNets and pre-activated ResNets, with less but still remarkable improvements for the VGGs. Except for VGGs, we can achieve a relatively 20$\%$ to $30\%$ testing error rate reduction across all neural nets. All results presented here and in the rest of this paper are the median of 5 independent trials. We also compare with SVM as an alternative output activation and observe that the performance are still inferior to the DNN-WNLL. Note that the bigger batch-size is to ensure the interpolation quality of the WNLL. A reasonable concern is that the performance increase comes from the variance reduction due to increasing the batch size. However, experiments were done with a batch size of $2000$ for vanilla networks deteriorates the test accuracy. 
\medskip

\begin{table*}[!ht]
\centering
\caption{Test errors of the vanilla DNNs, SVM and WNLL activated ones trained on the entire, the first $10$K, and the first $1$K instances of the training set of the CIFAR10 dataset. (Median of 5 independent trials)}\label{Table:Acc-List}
\label{Cifar10}
\begin{tabular}{cccccccc}
\toprule[1.0pt]
 Network &  \multicolumn{3}{c}{ Whole} &  \multicolumn{2}{c}{ 10000} & \multicolumn{2}{c}{ 1000}\\
\midrule[0.8pt]
{}  & {\small \textbf{ Vanilla}} &{\small \textbf{ WNLL}} &{\small \textbf{ SVM}} &{\small \textbf{ Vanilla}}   &{\small \textbf{ WNLL}}   &{\small \textbf{ Vanilla}}   &{\small \textbf{ WNLL}}  \\
{\small VGG11} 			& 9.23\% &  {\bf 7.35\%}  & 9.28\%  & 10.37\% & {\bf 8.88\%}   & 26.75\% & {\bf 24.10\%} \\
{\small VGG13} 			& 6.66\% &  {\bf 5.58\%}  & 7.47\%  & 9.12\% & {\bf 7.64\%}   & 24.85\% & {\bf 22.56\%} \\
{\small VGG16} 			& 6.72\% &  {\bf 5.69\%}  & 7.29\%  & 9.01\% & {\bf 7.54\%}   & 25.41\% & {\bf 22.23\%} \\
{\small VGG19} 			& 6.95\% &  {\bf 5.92\%}  & 7.99\%  & 9.62\% & {\bf 8.09\%}   & 25.70\% & {\bf 22.87\%} \\
{\small ResNet20} 		& 9.06\% &  {\bf 7.09\%}  & 9.60\%  & 12.83\%& {\bf 9.96\%}   & 34.90\% & {\bf 29.91\%} \\
{\small ResNet32} 		& 7.99\% &  {\bf 5.95\%}  & 8.73\%  & 11.18\%& {\bf 8.15\%}   & 33.41\% & {\bf 28.78\%} \\
{\small ResNet44} 		& 7.31\% &  {\bf 5.70\%}  & 8.67\%  & 10.66\%& {\bf 7.96\%}   & 34.58\% & {\bf 27.94\%} \\
{\small ResNet56} 		& 7.24\% &  {\bf 5.61\%}  & 8.58\%  &  9.83\%& {\bf 7.61\%}   & 37.83\% & {\bf 28.18\%} \\
{\small ResNet110} 		& 6.41\% &  {\bf 4.98\%}  & 8.06\%  &  8.91\%& {\bf 7.13\%}   & 42.94\% & {\bf 28.29\%} \\
{\small ResNet18} 		& 6.16\% &  {\bf 4.65\%}  & 6.00\%  & 8.26\% & {\bf 6.29\%}   & 27.02\% & {\bf 22.48\%} \\
{\small ResNet34} 		& 5.93\% &  {\bf 4.26\%}  & 6.32\%  & 8.31\% & {\bf 6.11\%}   & 26.47\% & {\bf 20.27\%} \\
{\small ResNet50} 		& 6.24\% &  {\bf 4.17\%}  & 6.63\%  & 9.64\% & {\bf 6.49\%}   & 29.69\% & {\bf 20.19\%} \\
{\small PreActResNet18}   & 6.21\% &  {\bf 4.74\%}  & 6.38\%  & 8.20\% & {\bf 6.61\%}   & 27.36\% & {\bf 21.88\%} \\
{\small PreActResNet34} 	& 6.08\% &  {\bf 4.40\%}  & 5.88\%  & 8.52\% & {\bf 6.34\%}   & 23.56\% & {\bf 19.02\%} \\
{\small PreActResNet50} 	& 6.05\% &  {\bf 4.27\%}  & 5.91\%  & 9.18\% & {\bf 6.05\%}   & 25.05\% & {\bf 18.61\%} \\
\bottomrule[1.0pt]
\end{tabular}
\end{table*}

\noindent We list the error rates of the $15$ different DNNs with either the softmax or the WNLL activation on the CIFAR10 and CIFAR100 in Tables~\ref{Cifar10} and \ref{Cifar100}, respectively.
On the CIFAR10, DNN-WNLL outperforms the vanilla ones with around 1.5$\%$ to 2.0$\%$ absolute, or 20$\%$ to 30$\%$ relative error rate reduction. The improvements on the CIFAR100 by using the WNLL activation are more remarkable than that on the CIFAR10. We independently ran the vanilla DNNs on both datasets, and our results are consistent with the original reports and other researchers' reproductions \cite{ResNet,IdentityMap:2016,DenseNet}. We provide experimental results of DNNs' performance on SVHN data in Table~\ref{SVHN-Whole}. Interestingly, the improvement is more significant on more challenge tasks which suggest a potential for our methods to succeed on other tasks/datasets.
\medskip

\begin{table*}[!ht]
\centering
\caption{Test errors of the vanilla DNNs v.s. the WNLL activated DNNs on the CIFAR100 dataset. (Median of 5 independent trials)}
\label{Cifar100}
\centering
\begin{tabular}{ccc}
\toprule[1.0pt]
\textbf{Network} & \textbf{Vanilla DNNs} & \textbf{WNLL DNNs}\\
\midrule[0.8pt]
VGG11           & 32.68\% & {\bf 28.80\%}  \\
VGG13           & 29.03\% & {\bf 25.21\%}  \\
VGG16           & 28.59\% & {\bf 25.72\%}  \\
VGG19           & 28.55\% & {\bf 25.07\%}  \\
ResNet20 		& 35.79\% & {\bf 31.53\%}  \\
ResNet32 		& 32.01\% & {\bf 28.04\%}  \\
ResNet44        & 31.07\% & {\bf 26.32\%}  \\
ResNet56        & 30.03\% & {\bf 25.36\%} \\
ResNet110       & 28.86\% & {\bf 23.74\%} \\
ResNet18 	     & 27.57\% & {\bf 22.89\%}\\
ResNet34 	     & 25.55\% & {\bf 20.78\%}\\
ResNet50        & 25.09\% & {\bf 20.45\%} \\
PreActResNet18  & 28.62\% & {\bf 23.45\%} \\
PreActResNet34  & 26.84\% & {\bf 21.97\%} \\
PreActResNet50  & 25.95\% & {\bf 21.51\%} \\
\bottomrule[1.0pt]
\end{tabular}
\end{table*}

\begin{table*}[!ht]
\centering
\caption{Test errors of the vanilla DNNs v.s. the WNLL activated DNNs on the SVHN dataset. (Median of 5 independent trials)}
\label{SVHN-Whole}
\centering
\begin{tabular}{ccc}
\toprule[1.0pt]
\textbf{Network} & \textbf{Vanilla DNNs} & \textbf{WNLL DNNs}\\
\midrule[0.8pt]
ResNet20        &  3.76\% & {\bf 3.44\%}\\
ResNet32        &  3.28\% & {\bf 2.96\%}\\
ResNet44        &  2.84\% & {\bf 2.56\%}\\
ResNet56        &  2.64\% & {\bf 2.32\%}\\
ResNet110       &  2.55\% & {\bf 2.26\%}\\
ResNet18 		&  3.96\% & {\bf 3.65\%}\\
ResNet34 		&  3.81\% & {\bf 3.54\%}\\
PreActResNet18 	&  4.03\% & {\bf 3.70\%}\\
PreActResNet34 	&  3.66\% & {\bf 3.32\%}\\
\bottomrule[1.0pt]
\end{tabular}
\end{table*}

\subsection{Adversarial Robustness}
We carry out experiments on the benchmark MNIST and CIFAR10 datasets to show the efficiency of using the graph interpolating activation for adversarial defense.
For MNIST, we train the Small-CNN that is used in \citep{Zhang:2019-Trades} by running $100$ epochs of PGD adversarial training with $\epsilon=0.3$, $\alpha=0.01$, and $M=40$. We let the initial learning rate be $0.1$ and decay by a factor of $10$ at the $50$th epoch. For CIFAR10, we consider three benchmark models: ResNet20, ResNet56, and WideResNet34. We train these models on the CIFAR10 dataset by running $120$ epochs of PGD adversarial training with $\epsilon=8/255$, $\alpha=2/255$, and $M=10$. The initial learning rate is set to be $0.1$ and decays by a factor of $10$ at the $80$th, $100$th, and $110$th epochs, respectively. After the robust models have been trained by the PGD adversarial training, we test their natural accuracies on the clean images and robust accuracies on the adversarial images crafted by attacking these robustly trained models by the aforementioned three adversarial attacks, where the parameters are set as follows
\begin{itemize}
\item {\bf FGSM:} In Eqs..~(\ref{FGSM}) and (\ref{FGSM-WNLL}), we let $\epsilon=8/255$ and $0.3$ to attack DNNs for CIFAR10 and MNIST classification, respectively.

\item {\bf IFGSM:} We denote the $n$-step IFGSM attack as IFGSM$^n$. To attack DNNs for CIFAR10 classification, we let $\epsilon=8/255$ and $\alpha=1/255$ in Eqs.~(\ref{IFGSM}) and (\ref{IFGSM-WNLL}) for both IFGSM$^{10}$ and IFGSM$^{20}$ attacks. For MNIST, we let $\epsilon=0.3$ and $\alpha=0.01$ in Eqs.~(\ref{IFGSM}) and (\ref{IFGSM-WNLL}) for IFGSM$^{40}$ and IFGSM$^{100}$ attacks.

\item {\bf C\&W:} For adversarial attack on the CIFAR10 dataset, we let $\kappa=0$ and $c=10$ in Eqs.~(\ref{CWL2}) and (\ref{CWL2-WNLL}), and we run $50$ iterations of the Adam optimizer with learning rate $0.006$ to find the optimal C\&W attack in $\ell_2$-norm on the clean images. To search for the optimal C\&W attack in $\ell_2$-norm on the MNIST data, we run $100$ iterations of the Adam optimizer with learning rate $0.003$ with $\kappa=0$ and $c=10$ in Eqs.~(\ref{CWL2}) and (\ref{CWL2-WNLL}).
\end{itemize}

\noindent We consider both white-box and black-box attacks. In the black-box attack, we apply the given adversarial attack to attack another oracle model in the white-box fashion, and then we use the target model to classify the adversarial images crafted by attacking the oracle model. 
\medskip

\begin{table}[!ht]
\centering
\fontsize{8.5}{8.5}\selectfont
\begin{threeparttable}
\caption{Natural and robust accuracies under different white-box adversarial attacks of different robustly trained models on the MNIST dataset.}\label{Table:White-Box-MNIST}
\begin{tabular}{cccccc}
\toprule[1.0pt]
Model  & $\mathcal{A}_{\rm nat}$ &$\mathcal{A}_{\rm rob}$ (FGSM) & $\mathcal{A}_{\rm rob}$ (IFGSM$^{40}$) &$\mathcal{A}_{\rm rob}$ (IFGSM$^{100}$)  & $\mathcal{A}_{\rm rob}$ (C\&W) \cr
\midrule[0.8pt]
Small-CNN           & 99.33\%  & 98.17\% & 96.27\%& 96.09\% & 95.31\%\cr
Small-CNN-WNLL      & 99.39\%  & 98.35\% & 97.36\%& 96.90\% & 97.55\%\cr
\bottomrule[1.0pt]
\end{tabular}
\end{threeparttable}
\end{table}

\begin{table}[!ht]
\centering
\fontsize{8.5}{8.5}\selectfont
\begin{threeparttable}
\caption{Robust accuracies under different black-box adversarial attacks of different robustly trained models on the MNIST dataset.}\label{Table:Black-Box-MNIST}
\begin{tabular}{cccccc}
\toprule[1.0pt]
Model & Oracle  & $\mathcal{A}_{\rm rob}$ (FGSM) & $\mathcal{A}_{\rm rob}$ (IFGSM$^{40}$) &$\mathcal{A}_{\rm rob}$ (IFGSM$^{100}$)  & $\mathcal{A}_{\rm rob}$ (C\&W) \cr
\midrule[0.8pt]
Small-CNN-WNLL   &Small-CNN   & 98.40\%  & 97.47\% & 97.40\%&  98.14\% \cr
\bottomrule[1.0pt]
\end{tabular}
\end{threeparttable}
\end{table}

\noindent Table~\ref{Table:White-Box-MNIST} lists both natural and robust accuracies of the PGD adversarially trained Small-CNN with either the softmax or the WNLL output activation function on the MNIST. Small-CNN with the WNLL activation is remarkably more accurate on both clean and adversarial images, e.g., for Small-CNN, the natural accuracies for the softmax and the WNLL activation functions are $99.33$\% and $99.39$\%, respectively. The robust accuracies for Small-CNN and Small-CNN-WNLL are $98.17$\% v.s. $98.35$\%, $96.27$\% v.s. $97.36$\%, $96.09$\% v.s. $96.90$\%, and $95.31$\% v.s. $97.55$\%, respectively, to the FGSM, IFGSM$^{40}$, IFGSM$^{100}$, and C\&W attacks in the white-box scenario. We regard Small-CNN as the oracle model to perform black-box attacks on the Small-CNN-WNLL, the corresponding robust accuracies to the above four adversarial attacks are listed in Table~\ref{Table:Black-Box-MNIST}. In the MNIST experiment, black-box attacks are less effective than the white-box attacks.
\medskip

\begin{table}[!ht]
\centering
\fontsize{8.5}{8.5}\selectfont
\begin{threeparttable}
\caption{Natural and robust accuracies under different white-box adversarial attacks of different robustly trained models on the CIFAR10 dataset.}\label{Table:White-Box-Cifar10}
\begin{tabular}{cccccc}
\toprule[1.0pt]
Model  & $\mathcal{A}_{\rm nat}$ &$\mathcal{A}_{\rm rob}$ (FGSM) & $\mathcal{A}_{\rm rob}$ (IFGSM$^{20}$) &$\mathcal{A}_{\rm rob}$ (IFGSM$^{100}$)  & $\mathcal{A}_{\rm rob}$ (C\&W) \cr
\midrule[0.8pt]
ResNet20               &  75.11\% & 50.89\% & 46.03\% & 46.01\% & 58.73\%\cr
ResNet20-WNLL          &  75.53\% & 55.76\% & 53.31\% & 53.26\% & 63.82\%\cr
ResNet56               &  79.32\% & 55.05\% & 50.98\% & 50.06\% & 61.75\%\cr
ResNet56-WNLL          &  79.52\% & 60.50\% & 58.19\% & 57.26\% & 67.93\%\cr
WideResNet34           &  84.05\% & 51.93\% & 48.93\% & 48.32\% & 59.04\%\cr
WideResNet34-WNLL      &  84.95\% & 65.50\% & 63.03\% & 62.25\% & 72.37\%\cr
\bottomrule[1.0pt]
\end{tabular}
\end{threeparttable}
\end{table}

\begin{table}[!ht]
\centering
\fontsize{7.5}{7.5}\selectfont
\begin{threeparttable}
\caption{Robust accuracies under different black-box adversarial attacks of different robustly trained models on the CIFAR10.}\label{Table:Black-Box-Cifar10}
\begin{tabular}{cccccc}
\toprule[1.0pt]
Model & Oracle  & $\mathcal{A}_{\rm rob}$ (FGSM) & $\mathcal{A}_{\rm rob}$ (IFGSM$^{20}$) &$\mathcal{A}_{\rm rob}$ (IFGSM$^{100}$)  & $\mathcal{A}_{\rm rob}$ (C\&W) \cr
\midrule[0.8pt]
ResNet20-WNLL     & ResNet20         & 55.91\% & 53.44\%& 53.35\% & 65.13\%\cr
ResNet56-WNLL     & ResNet56         & 60.00\% & 57.94\%& 57.85\% & 70.47\%\cr
WideResNet34-WNLL & WideResNet34     & 67.19\% & 67.07\%& 67.17\% & 81.17\%\cr
\bottomrule[1.0pt]
\end{tabular}
\end{threeparttable}
\end{table}

\noindent Next, we consider the adversarial defense capability of DNNs with the WNLL activation on the CIFAR10 dataset.
Table~\ref{Table:White-Box-Cifar10} lists the natural and robust accuracies, under the white-box attacks, of the standard ResNet20, ResNet56, and WideResNet34-10 and their counterpart with the WNLL activation. These results show that the robustly trained ResNets with the WNLL activation slightly improves natural accuracies on the clean images, while the robust accuracies are significantly improved. For instance, under the FGSM and C\&W attacks, the WNLL activation can boost robust accuracy by $\sim 5$\%; and under the IFGSM$^{20}$ and IFGSM$^{40}$ attacks, the robust accuracy improvement is up to $\sim 7$\%. For the WideResNet34-10, under the IFGSM$^{20}$ attack, we achieve accuracy $63.03$\% which outperforms the results of \cite{Zhang:2019-Trades} ($56.61$\%) by more than $6$\%. For black-box attacks on DNNs with the WNLL activation, we regard the counterpart DNNs with the softmax activation as the oracle models. The robust accuracies of ResNet20-WNLL, ResNet56-WNLL, and WideResNet34-10-WNLL are listed in Table~\ref{Table:Black-Box-Cifar10}. Again, the black-box attacks are less effective than the white-box ones.
\medskip

\begin{table}[!ht]
\centering
\fontsize{8.5}{8.5}\selectfont
\begin{threeparttable}
\caption{Natural and robust accuracies under different white-box adversarial attacks of ResNet56-WNLL with different number of points, in the form $(m, n)$, used for interpolation on the CIFAR10 dataset.}\label{Table:White-Box-Cifar10-WNLL-new}
\begin{tabular}{cccccc}
\toprule[1.0pt]
Model  & $\mathcal{A}_{\rm nat}$ &$\mathcal{A}_{\rm rob}$ (FGSM) & $\mathcal{A}_{\rm rob}$ (IFGSM$^{20}$) &$\mathcal{A}_{\rm rob}$ (IFGSM$^{100}$)  & $\mathcal{A}_{\rm rob}$ (C\&W) \cr
\midrule[0.8pt]
ResNet56-WNLL (15, 8)         &  79.89\% & 59.71\% & 57.85\%& 56.53\% & 67.91\%\cr 
ResNet56-WNLL (30, 15)        &  79.52\% & 60.50\% & 58.19\%& 57.26\% & 67.93\%\cr
ResNet56-WNLL (45, 23)        &  78.92\% & 59.50\% & 57.94\%& 57.06\% & 66.26\%\cr 
ResNet56-WNLL (60, 30)        &  77.92\% & 58.04\% & 55.80\%& 54.97\% & 67.74\%\cr 
\bottomrule[1.0pt]
\end{tabular}
\end{threeparttable}
\end{table}

\noindent Furthermore, we consider the influence of the number of nearest neighbors $m$ with the $n$th-nearest neighbor used to normalize the weights in Eq.~(\ref{WNLL}) in the WNLL interpolation. Table~\ref{Table:White-Box-Cifar10-WNLL-new} lists the natural and robust accuracies of the ResNet56-WNLL with the different number of nearest neighbors, $(m, n)$, involved in the WNLL interpolation. The natural accuracy decays as a greater number of nearest neighbors are used for interpolation, and the robust accuracies are maximized when $(m, n)=(30, 15)$. When more nearest neighbors are used for interpolation, the robust accuracies decay. This issue might be due to the fact that these nearest neighbors are only selected from a finite number of data points and the resulted nearest neighbors are far from the real nearest neighbors.
\medskip

\begin{figure}[!ht]
\centering
\begin{tabular}{cc}
\includegraphics[width=0.45\columnwidth]{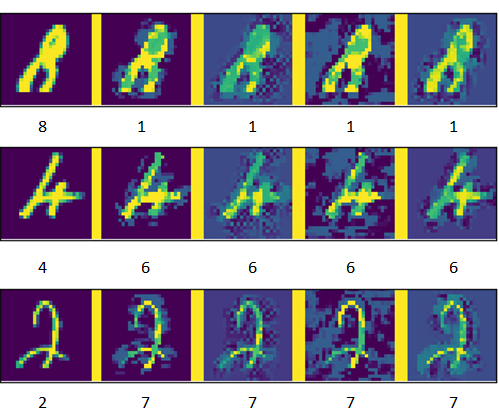}&
\includegraphics[width=0.46\columnwidth]{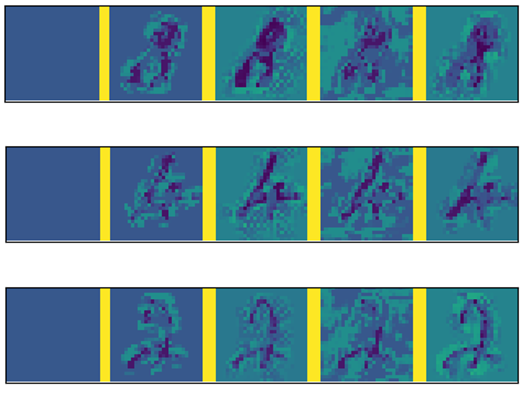}\\
Adversarial Images & Adversarial Noise\\
\end{tabular}
\caption{Adversarial images (left panel) selected from the MNIST dataset and the corresponding adversarial noise (right panel). Column 1: cleaning image and noise (no noise in this case); Column 2-3: adversarial images and noise crafted by IFGSM$^{40}$ and C\&W attacks on the small CNN, respectively; Column 4-5: adversarial images and noise crafted by IFGSM$^{40}$ and C\&W attacks on the small CNN-WNLL, respectively. The predicted labels for the adversarial images are listed below the adversarial images in the left panel.}
\label{fig-Attack-MNIST}
\end{figure}

\begin{figure}[!ht]
\centering
\begin{tabular}{cc}
\includegraphics[width=0.45\columnwidth]{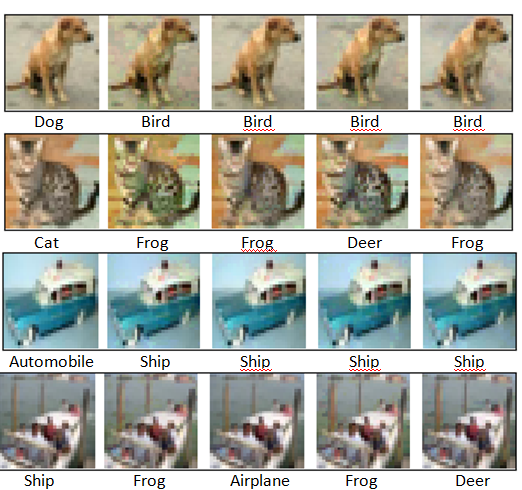}&
\includegraphics[width=0.455\columnwidth]{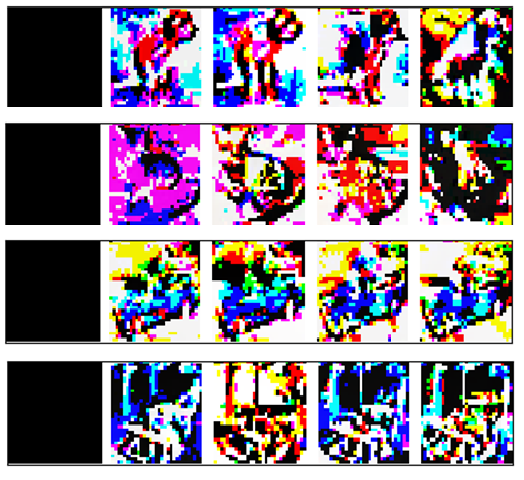}\\
Adversarial Images & Adversarial Noise\\
\end{tabular}
\caption{Adversarial images (left panel) selected from the CIFAR10 dataset and the corresponding adversarial noise (right panel). Column 1: cleaning image and noise (no noise in this case); Column 2-3: adversarial images and noise crafted by IFGSM$^{40}$ and C\&W attacks on the ResNet20, respectively; Column 4-5: adversarial images and noise crafted by IFGSM$^{20}$ and C\&W attacks on the ResNet20-WNLL, respectively. The predicted labels for the adversarial images are listed below the adversarial images in the left panel.}
\label{fig-Attack-CIFAR10}
\end{figure}

\noindent Finally, let us look at the adversarial images and the adversarial noise crafted by adversarial attacks on DNNs with both softmax and WNLL activation functions. Figures~\ref{fig-Attack-MNIST} and \ref{fig-Attack-CIFAR10} depict adversarial images and adversarial noise of the MNIST and the CIFAR10 obtained by applying different adversarial attacks to Small-CNN and ResNet20 with both softmax and WNLL activation functions. All these adversarial images are misclassified by DNNs with both the softmax and the WILL activation. However, they can be easily classified by human-beings.

\subsection{Semi-supervised Learning}
In this subsection, we apply the DNN-WNLL to semi-supervised learning where we have access to all the training data of the CIFAR10 but only part of them are labeled. We can use the unlabeled data to build a graph for the WNLL interpolation in semi-supervised learning, while not in data-efficient learning. We list the accuracies of the semi-supervised learning when $1$K and $10$K training data are labeled to train DNNs in Table~\ref{SemiSupervised-I}. Compared to the results in Table~\ref{Table:Acc-List},  semi-supervised learning has better accuracy with the same number of labeled training data.

\begin{table*}[!ht]
\centering
\caption{Test error of DNNs with the WNLL output activation for the CIFAR10 classification in the semi-supervised learning setting.}
\label{SemiSupervised-I}
\begin{tabular}{ccc}
\toprule
Network &  1K (Labeled)/49K (Unlabeled) &  10K (Labeled)/ 40K (Unlabeled)\\
\midrule
ResNet20-WNLL 		& 27.02\% & 9.01\%  \\
ResNet32-WNLL 		& 26.28\% & 7.53\%  \\
ResNet44-WNLL 		& 25.63\% & 7.25\%  \\
ResNet56-WNLL 		& 25.53\% & 6.99\%  \\
ResNet110-WNLL 	    & 25.38\% & 6.50\%  \\
\bottomrule
\end{tabular}
\end{table*}

\section{Geometric Explanations}\label{Section:Explanation}
In this section, we will consider the representations learned by DNNs with two different output activation functions. As an illustration, we randomly select $1000$ training instances and $100$ testing data each for the airplane and automobile classes from the CIFAR10 dataset. We consider two different strategies to visualize the features learned by ResNet56 and ResNet56-WNLL for the above randomly selected data.
\begin{itemize}
    \item {\bf Strategy I:} Apply the principal component analysis (PCA) to reduce the $64$D features output right before the softmax/WNLL activation to $2$D. 
    
    \item {\bf Strategy II:} Add an additional fully connected layer before the output activation function. This fully connected layer will help to learn the $2$D representations.
\end{itemize}

\begin{figure}[!ht]
\centering
\begin{tabular}{cc}
\includegraphics[width=0.46\columnwidth]{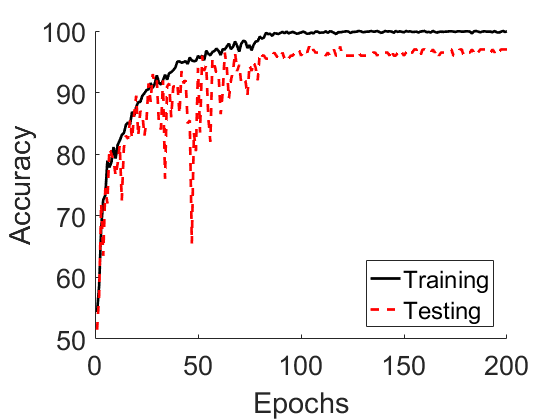} &
\includegraphics[width=0.46\columnwidth]{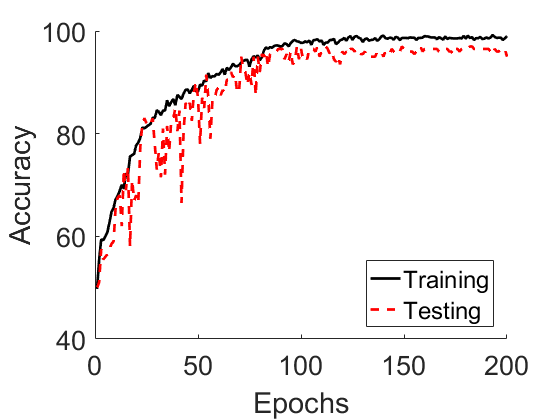}\\
(a)&(b)\\
\end{tabular}
\caption{Epochs v.s. accuracy in training ResNet56 on the CIFAR10. (a): without the additional FC layer; (b): with the additional FC layer.}
\label{fig:Add2x2}
\end{figure}

\noindent We first show that in {\bf Strategy II}, the newly added fully connected (FC) layer does not affect the performance of the original ResNet56 much. We train and test the ResNet56 with and without the additional FC layer on the aforementioned randomly selected training and testing data. As shown in Fig.~\ref{fig:Add2x2}, the training and testing accuracies evolution are essentially the same for ResNet56 with and without the additional FC layer.

\subsection{Improving Generalization}
\begin{figure}[!ht]
\centering
\begin{tabular}{cc}
\includegraphics[width=0.46\columnwidth]{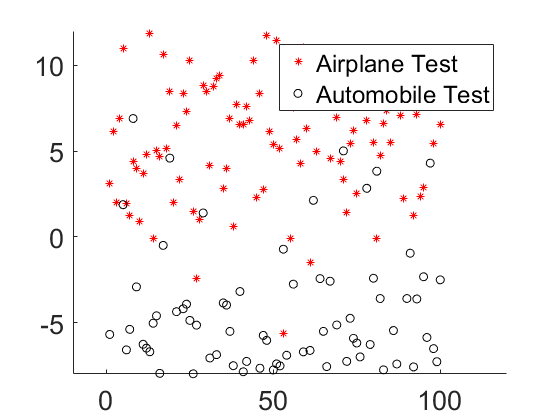}&
\includegraphics[width=0.46\columnwidth]{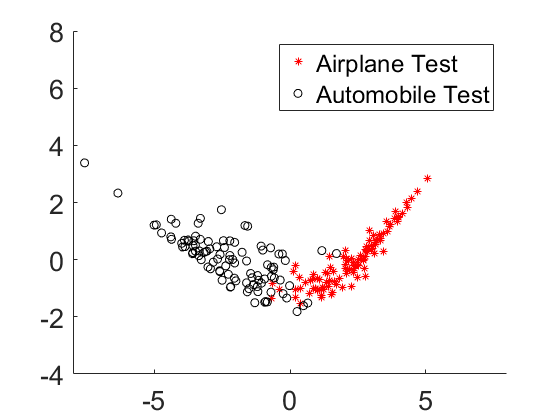}\\
(a)&(b)\\
\includegraphics[width=0.46\columnwidth]{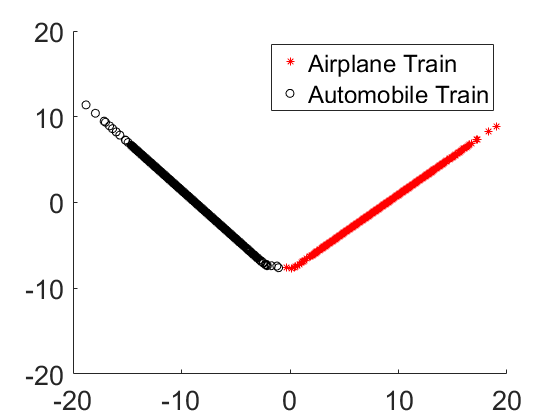}&
\includegraphics[width=0.46\columnwidth]{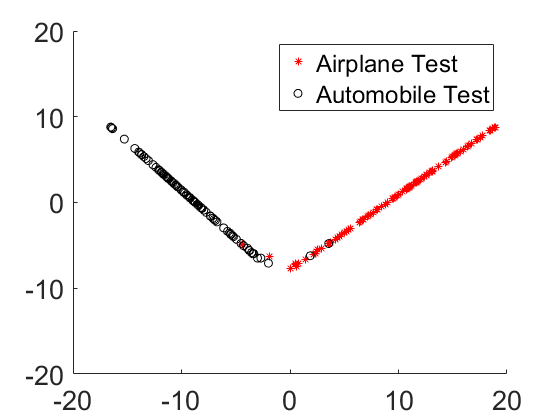}\\
(c)&(c)\\
\end{tabular}
\caption{
Visualization of the features learned by ResNet56 with the softmax ((a), (b)) and the WNLL ((c), (d)) activation functions. 
(a): the 2D features of the airplane and automobile data in the test set learned by the ResNet56 with an additional $2\times 2$ linear layer; (b): the first two principal components of the features of the airplane and automobile data in the test set learned by the ResNet56; (c) and (d) plot the first two principal components of features of the airplane and automobile data in the training and test set learned by the ResNet56-WNLL. All experiments are done on the CIFAR10 dataset.}
\label{fig:FeaturesGeometry-Clean}
\end{figure}

Figure~\ref{fig:FeaturesGeometry-Clean} plots the representations for the selected airplane and automobile data from the CIFAR10 dataset. First, panels (a) and (b) show the features of the test set learned by ResNet56 visualized by the proposed two strategies. In both cases, the features are well separated, in general, with a small overlapping which causes some misclassification. Charts (c) and (d) depict the first two principal components (PCs) learned by ResNet56-WNLL for the selected training and testing data. The PCs of the features learned by ResNet56-WILL is better separated than that of ResNet56's (Fig.~\ref{fig:FeaturesGeometry-Clean}), and it indicates that ResNet56-WILL is more accurate in classifying the randomly selected data.

\subsection{Improving Adversarial Robustness}
\begin{figure}[!ht]
\centering
\begin{tabular}{cc}
\includegraphics[width=0.46\columnwidth]{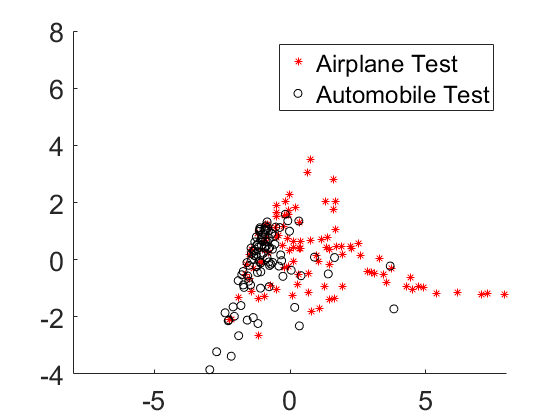}&
\includegraphics[width=0.46\columnwidth]{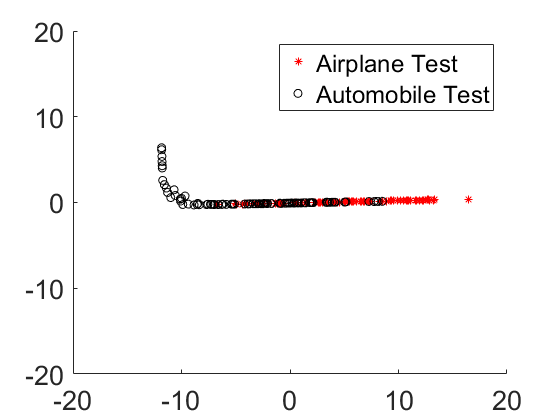}\\
(a)&(b)\\
\end{tabular}
\caption{Visualization of the first two principal components of the adversarial images' (IFGSM$^1$ attack) features learned by ResNet56 with the softmax (a) and the WNLL (b) activation functions, respectively.}
\label{fig:FeaturesGeometry}
\end{figure}

First, let us look at how the adversarial attack changes the geometry of the learned representations. We consider the simple one-step IFGSM attack, IFGSM$^1$, with the same parameters used before. Figure~\ref{fig:FeaturesGeometry} shows the first two PCs of the representations learned by ResNet56 and ResNet56-WNLL for the adversarial test images. These PCs show that the adversarial attack makes the features of the two different classes mixed and therefore drastically reduces the classification accuracy.
\medskip

\begin{figure}[!ht]
\centering
\begin{tabular}{cc}
\includegraphics[width=0.45\columnwidth]{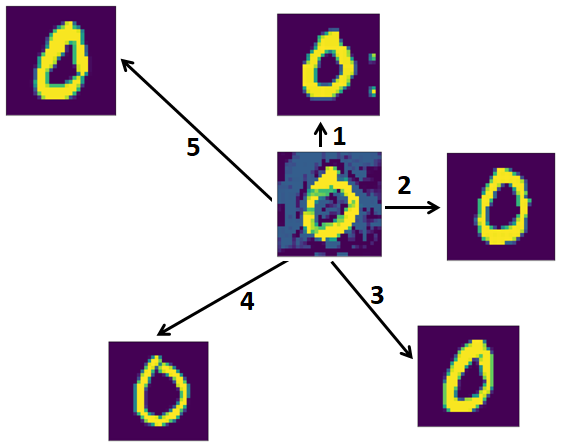}&
\includegraphics[width=0.45\columnwidth]{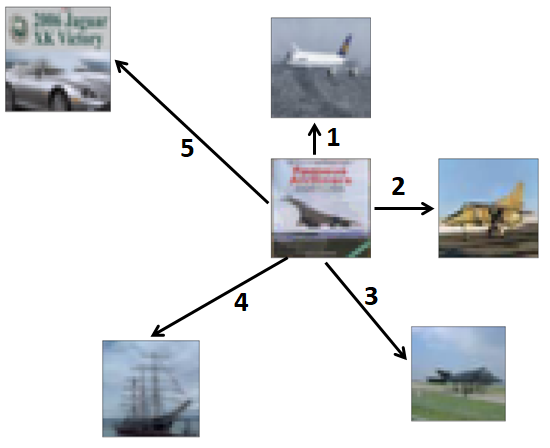}\\
MNIST & CIFAR10\\
\end{tabular}
\caption{A randomly selected adversarial image and their top five nearest neighbors in the clean training set searched based on the distance between features output from the layer before the WNLL activation layer. Left: IFGSM$^{40}$ attack on the small CNN-WNLL; Right: IFGSM$^{20}$ attack on the ResNet20-WNLL.}
\label{fig-NNB}
\end{figure}

\noindent Second, we consider how the WNLL interpolation helps to improve adversarial robustness. We randomly pick up an adversarial image that is misclassified by the standard DNNs with the softmax activation from the MNIST and the CIFAR10, respectively. The top five nearest neighbors in the deep feature space from the clean training data, of these two adversarial images, are shown in Fig.~\ref{fig-NNB}. For the MNIST digit, all the nearest neighbors belong to the same class as the adversarial image; and for the CIFAR10 adversarial image, the top three neighbors belong to the same category as the adversarial one. These nearest neighbors will guide DNN-WNLL to classify the adversarial images correctly.

\section{Concluding Remarks}\label{Section:Conclusion}
In this paper, we leveraged ideas from the manifold learning and proposed to replace the output activation function of the conventional deep neural nets (DNNs), typically the softmax function, with a graph Laplacian-based high dimensional interpolating function. This simple modification is applicable to any of the existing off-the-shelf DNNs with the softmax activation enables DNNs to make sufficient use of the manifold structure of data. Furthermore, we developed end-to-end and multi-staged training and testing algorithms for the proposed DNN with the interpolating function as its output activation. On the one hand, the proposed new framework remarkably improves both generalizability and robustness of the baseline DNNs; on the other hand, the new framework is suitable for data-efficient machine learning. These improvements are consistent across networks of different types and with a different number of layers. The increase in generalization accuracy could also be used to train smaller models with the same accuracy, which has great potential for mobile device applications.
\medskip

\noindent In this work, we utilized a special kind of graph interpolating function as DNNs' output activation. An alternative approach is to learn such an interpolating function instead of using one which is fixed. This approach is under our consideration.


\section*{Acknowledgments}
This material is based on research sponsored by the Air Force Research Laboratory under grant numbers FA9550-18-0167 and MURI FA9550-18-1-0502, the Office of Naval Research under grant number N00014-18-1-2527, the U.S. Department of Energy under grant number DOE SC0013838, and by the National Science Foundation under grant number DMS-1554564 (STROBE).


\end{document}